%% file: main.tex
\documentclass{article}

\usepackage{microtype}
\usepackage{graphicx}
\usepackage{booktabs}
\usepackage{makecell}
\usepackage{threeparttable}
\usepackage{amsmath}
\usepackage{amssymb}
\usepackage{amsthm}
\usepackage{subcaption}

\usepackage{hyperref}

\usepackage[accepted,preprint]{mlsys_preprint}
\usepackage[capitalize,noabbrev]{cleveref}

\DeclareMathOperator*{\argmax}{arg\,max}
\newtheorem{lemma}{Lemma}

\date{March 2026}

\begin{document}

\twocolumn[
\mlsystitle{
Template Gap Reconstruction for Shell Output Simulation with Small Language Models
}

\begin{mlsysauthorlist}
\mlsysauthor{Jarrod Ragsdale}{utsa}
\mlsysauthor{Rajendra Boppana}{utsa}
\end{mlsysauthorlist}

\mlsysaffiliation{utsa}{Department of Computer Science, University of Texas at San Antonio, USA}
\mlsyscorrespondingauthor{Jarrod Ragsdale}{jarrod.ragsdale@my.utsa.edu}

\mlsyskeywords{Machine Learning, MLSys}

\vskip 0.3in

\input{content/0_abstract}
]

\printAffiliationsAndNotice{}

\input{content/1_intro_v2}
\input{content/2_background_v2}
\input{content/3_method_v2}
\input{content/4_eval_v2}
\input{content/5_discussion}
\input{content/6_conclusion}

\section*{Acknowledgements}
The research was sponsored by the Army Research Laboratory, accomplished under Cooperative Agreement Number W911NF-24-2-0180, and conducted in the Systems and Networks (SyN) Lab at The University of Texas at San Antonio using the Cyber Deception Testbed, funded by the Army Research Office under grant number W911NF-21-1-0188 and the National Security Agency through the awards H98230-21-1-0171 and H98230-20-1-0392. The views and conclusions contained in this document are those of the authors and should not be interpreted as representing the official policies, either expressed or implied, of the Army Research Laboratory or the U.S. Government. Parts of this research were facilitated by the use of experimental hardware via the Nvidia Launchpad program.

\bibliography{references}
\bibliographystyle{mlsys2025}

\clearpage
\onecolumn
\raggedbottom
\appendix
\input{content/7_appendix}
\input{content/73_append}

\end{document}

%% file: content/0_abstract.tex
\begin{abstract}
Small language models offer a compact basis for shell simulation without executing submitted commands, but full-output decoding requires them to predict recurring utility structure alongside content that varies across executions.
We introduce template gap reconstruction (TGR), which supplies reusable output literals interspersed with sentinels to be interleaved with autoregressively generated gaps.
A template selector chooses which template to be reconstructed, with the empty template retaining full-output decoding.
We construct a template bank from 18.7 million outputs collected across heterogeneous execution environments and train a template selector and five reconstruction models initialized from various small language model checkpoints on a target environment excluded from bank construction.
Evaluation demonstrates that TGR improves output edit similarity (O-ES) over full-output decoding (FOD) when the optimal template covers more than 25\% of the template, while performing similarly for lower-coverage outputs.
Additionally, TGR outperforms FOD for shorter gap sequences, with O-ES values of 0.964 and 0.812, respectively, while maintaining similar performance for longer sequences.
TGR reduces the number of tokens generated by 26.3--38.3\% across used tokenizer families.
These results support using a reusable output structure as a practical design choice for small shell simulators, improving response fidelity and shortening generation targets while keeping template selection and residual prediction central to response quality.
\end{abstract}

%% file: content/1_intro_v2.tex
\section{Introduction}
\label{sec:intro}
Small language models generating the output of provided shell inputs offer a promising basis for command-line interaction under limited inference resources.
Such an execution-free interface supports compact learned honeypots that interact with untrusted users \cite{sladic2024llm}.
Prior work uses in-context learning with API-accessible large language models \cite{sladic2024llm} and fine-tuned locally deployed large language models to generate shell output \cite{otal2024llm,wei2026mantis,sladic2026improving}.
\citet{ragsdale2025shioenv} demonstrate that this task can be extended to smaller language models ($<$1B parameters) when trained on sufficient quantities of shell input-output pairs.
The success of these methods depends on the similarity of the generated output to that of a real system \cite{bridges2026sok}.
This paper is motivated in the improvement of shell output modeling accuracy using small lanuage models.

Inference depends on the text processed and generated, with autoregressive decoding producing each output token in sequence \cite{pope2023efficiently}.
Full-output decoding (FOD) requires a shell output model to predict filenames, values, diagnostics, punctuation, and internal whitespace alongside recurring headings, separators, and wording.
Supplying this recurring structure as context offers a way to reduce the prediction burden.

\begin{figure*}[htbp]
    \centering
    \includegraphics[width=0.98\linewidth]{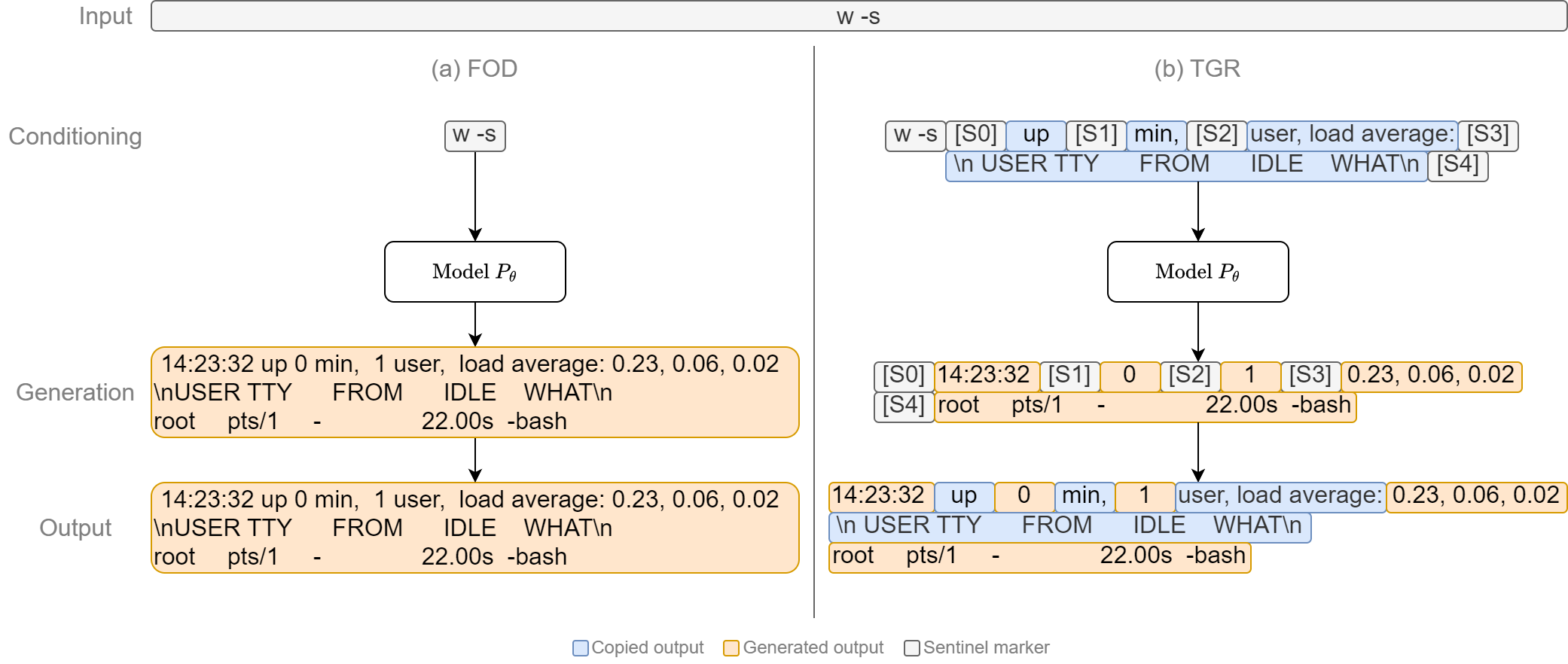}
    \caption{Full-output decoding (FOD) and template gap reconstruction (TGR) producing the same complete shell response. Blue marks copied template literals, and orange marks generated text. Sentinel markers delimit the generated gaps.}
    \label{fig:decoding-interfaces}
\end{figure*}

We investigate environment-invariant output templates: ordered literal spans shared by groups of recorded executions.
Greedy alignment extracts spans whose empirical invariance is scoped to the construction group.
Template gap reconstruction (TGR) supplies a template alongside the input, generates sentinel-delimited template gaps, and interleaves them with copied literals.
Retained text thus provides context without remaining in the generation target (\Cref{fig:decoding-interfaces}).

Before the output is known, an input-only selector with a shared Transformer encoder and command-specific heads chooses from a template bank containing the empty template $T_0$ for each command.
A nonempty selection invokes TGR, while selecting $T_0$ invokes ordinary FOD through the same reconstruction checkpoint.
The constructed template bank contains templates extracted from 18.7 million executions across heterogeneous environments.

We evaluate output similarity across five models fine-tuned for shell modeling and TGR spanning approximately 60--600 million parameters: CodeT5-small and CodeT5-base, ByT5-small and ByT5-base, and Qwen3-0.6B \cite{wang2021codet5,xue2022byt5,yang2025qwen3}.
A target environment supplies 1 million examples for selector and reconstruction training.
Oracle assignment uses the reference output to maximize admissible coverage and defines the coverage and reference-gap length used to group all methods.

Selected TGR improves mean complete-response fidelity across all five models.
Oracle TGR improves mean output edit similarity (O-ES) over FOD in every reported coverage group above 25\%, with learned selection yielding consistent gains above 50\%.
Selected TGR achieves O-ES of 0.879 compared with 0.742 for FOD in the $(50,75]\%$ coverage group.
For responses with 0--64 reference-gap characters, equally weighting the three reported length groups gives mean O-ES of 0.964 for oracle TGR compared with 0.812 for FOD.
The advantage diminishes for longer gaps, with closer scores of 0.772 and 0.790, respectively, in the longest reference-gap group.
TGR also shortens mean tokenized reference targets by 26.3--38.3\% across tokenizer families, including sentinel overhead.

We make the following contributions:
\begin{enumerate}
    \item We formalize reusable templates for complete shell responses and provide a greedy construction procedure that preserves admissibility within each group of recorded executions.
    \item We combine an input-only, command-specific selector with a reconstruction checkpoint jointly trained for sentinel-delimited gap generation and ordinary full-output fallback.
    \item We evaluate this interface for small language models at fixed reconstruction weights, distinguishing complete-response fidelity, generated-gap fidelity, and reference-target length across five backbones.
\end{enumerate}

%% file: content/2_background_v2.tex
\section{Background and Related Work}
\label{sec:background}

\paragraph{Small language models for shell simulation.}
Execution-derived specialization learns response behavior for a target environment \cite{ragsdale2025shioenv}.
The evaluated backbones include encoder--decoder CodeT5-small/base (approximately 60M/220M parameters) \cite{wang2021codet5}, byte-level encoder--decoder ByT5-small/base (300M/580M) \cite{xue2022byt5}, and decoder-only Qwen3-0.6B (600M) \cite{yang2025qwen3}.
At fixed numerical precision, smaller models require less weight storage, with their inference time and working memory depending on architecture, sequence lengths, and hardware \cite{pope2023efficiently}.
Ordinary autoregressive decoding invokes the model for successive output tokens, making target length another relevant design choice.

Shell responses depend on filesystem contents, permissions, variable bindings, and available programs \cite{posix2024shell,greenberg2019executable}.
Honeypots motivate shell-response generation without executing attacker-supplied commands on the host \cite{otal2024llm}.
Cowrie's emulation mode combines command handling with a fake filesystem and session logging \cite{cowrie}, while shelLM and advancedShelLM generate context-aware responses with language models \cite{sladic2024llm,sladic2026advancedshellm}.
SBASH examines system-prompt conditioning and retrieval augmentation for local Llama and Gemma models \cite{adebimpe2025sbash}.

Fine-tuning also supports local shell-response models.
Otal and Canbaz adapt Llama~3~8B using command--response supervision from Cowrie \cite{otal2024llm}.
MANTIS fine-tunes a Llama-3.2-3B-Instruct response agent on recorded Linux command--output pairs and combines it with command routing, filesystem management, and interaction history \cite{wei2026mantis}.
Sladic et al. compare supervised fine-tuning and prompting for seven local open-weight large language models, finding that adaptation benefits depend on training-data coverage \cite{sladic2026improving}.
ShIOEnv synthesizes shell inputs and records observed execution outputs at scale \cite{ragsdale2025shioenv}.
These records support both specialization and the identification of recurring response fragments across environments.

\paragraph{Template induction.}
Log parsers similarly separate recurring message structure from variable fields.
Spell incrementally compares token sequences using a longest common subsequence and updates the template associated with a log group \cite{du2016spell}.
Drain uses a fixed-depth tree to narrow candidate groups before matching messages and updating their templates \cite{he2017drain}.
Sequence alignment also supports pattern induction for generation: Barzilay and Lee align sentences from comparable corpora to construct paraphrasing patterns represented by word lattices \cite{barzilay2003learning}.

The presented procedure applies alignment to complete recorded shell responses, including multiline output.
Canonicalized inputs and exact observed exit codes define groups before induction.
Each template retains ordered, contiguous character spans whose boundaries identify positions for text.
The construction guarantee concerns their occurrence in every output of the observed group.
Using that structure for a new execution requires an input-conditioned decision about which template is appropriate.

\paragraph{Span infilling and template-based generation.}
T5's span-corruption objective replaces sampled spans with unique sentinels and trains the decoder to predict the removed text \cite{raffel2020t5}.
Donahue et al. train language models to fill arbitrary blanks using sequences that concatenate masked context and missing text \cite{donahue2020enabling}.
These interfaces expose retained text to the model while restricting the prediction target to missing spans.

Template-based generation makes output organization explicit.
Wiseman et al. learn latent discrete templates jointly with generation through a hidden semi-Markov decoder \cite{wiseman2018templates}.
TempLM constructs an explicit template-based generator from a pretrained language model by replacing data values in generated outputs with field slots and refining the templates \cite{zhang2023templm}.
TGR obtains its literals from repeated executions and uses their positions to similarly define sentinel-delimited targets.
Its gaps contain all text outside the retained literals, including changing values and stable text that induction did not preserve.
Gap scoring separately evaluates this generated text, whose length and content depend on the template.

\paragraph{Editing and constrained generation.}
Editing systems reuse text while predicting additions needed to complete an output.
EdiT5 combines keep/delete tagging and token reordering with autoregressive insertion of missing spans \cite{mallinson2022edit5}.
POINTER preserves supplied tokens while progressively inserting new text between them \cite{zhang2020pointer}.
Lexically constrained decoding incorporates required words or phrases into a complete hypothesis during search \cite{post2018constraineddecoding}.
TGR combines execution-derived literals with input-only selection from a command-specific bank.
The empty template invokes ordinary FOD in the same jointly trained checkpoint.
Admissible templates can leave gaps with different lengths, segmentation, and predictability, making their utility depend on both compatibility and residual prediction.

%% file: content/3_method_v2.tex
\section{Environment-Invariant Output Templates}
\label{sec:5_method}
Template gap reconstruction (TGR) predicts a shell response by combining copied template literals with generated residual spans.
The method comprises template construction, template-conditioned gap generation, and input-conditioned template selection.
Let $x$ denote a submitted shell input and $y$ its recorded output.

\subsection{Template Construction}
\label{sec:method_extraction}
Let $\Sigma$ be the character alphabet of the recorded output, including retained whitespace and control characters.
A template is an ordered sequence of non-empty text segments,
\[
T=(\tau_1,\ldots,\tau_k),\qquad \tau_i\in\Sigma^+.
\]
A template overlays $y\in\Sigma^*$, written $T\preceq y$, if there exist gaps $s_0,\ldots,s_k\in\Sigma^*$ such that
\begin{equation}
\label{eq:template-decomposition}
y=s_0\tau_1s_1\tau_2\cdots\tau_ks_k.
\end{equation}
Each template segment occurs in order and without overlap.
The gaps collect all remaining characters, including any leading or trailing text, and may be empty.
Template identity includes the literal text and its segment boundaries: each boundary determines a separate position for a gap.
For example, $(\texttt{ab},\texttt{cd})$ overlays \texttt{abXcd}, whereas $(\texttt{abcd})$ does not, although both templates retain four characters.
The empty template $T_0=()$ overlays every output, with $s_0=y$.
Admissibility is checked by matching each literal at its earliest occurrence after the preceding match.
An earlier match leaves at least as much suffix for later literals, so greedy failure implies that no overlay exists.

We group execution records by their canonicalized input and observed exit code.
Canonicalization abstracts environment-specific resource arguments while preserving the command and its argument structure.
For example, \texttt{ls -la} inputs with different directory arguments share the canonical form \texttt{ls -la [Path]}.
A template is admissible if it overlays every output in its group.

\begin{algorithm}[htbp]
\caption{Greedy admissibility-preserving template refinement}
\label{alg:template-extraction}
\begin{algorithmic}[1]
\REQUIRE Construction group $(y_1,\ldots,y_n)$
\ENSURE Template $T \preceq y$ for all $ y \in (y_1,\ldots,y_n)$
\IF{$n=0$ \OR $y_1=\epsilon$}
    \STATE \textbf{return} $()$
\ENDIF
\STATE $T\gets(y_1)$
\FOR{$i\gets 2$ \textbf{to} $n$}
    \STATE $y\gets y_i$; $T'\gets()$; $p\gets 1$
    \FOR{$j\gets 1$ \textbf{to} $|T|$}
        \IF{$p\leq |y|$}
            \STATE $B\gets\operatorname{LCSBlocks}(T_j,y_p\cdots y_{|y|})$
            \FOR{$h\gets 1$ \textbf{to} $|B|$}
                \STATE $\beta\gets B_h$
                \STATE $q\gets\operatorname{FirstOccurrence}(y,\beta,p)$
                \STATE $\operatorname{append}(T',\beta)$
                \STATE $p\gets q+|\beta|$
            \ENDFOR
        \ENDIF
    \ENDFOR
    \STATE $T\gets T'$
\ENDFOR
\STATE \textbf{return} $T$
\end{algorithmic}
\end{algorithm}

\paragraph{Coverage and refinement.}
Literal coverage measures the number of output characters retained by a template:
\begin{equation}
\label{eq:literal-coverage}
\operatorname{cov}(T)=\sum_{i=1}^{k}|\tau_i|.
\end{equation}
Concatenating admissible literals gives a common subsequence of the construction outputs.
Maximum admissible coverage therefore corresponds to a longest common subsequence (LCS) across the outputs.
The decision version for an arbitrary number of sequences is NP-complete \cite{maier1978complexity}, motivating the greedy procedure in \Cref{alg:template-extraction}.

The first output initializes a single-segment template.
For each subsequent output, the algorithm traverses the literals in order, aligning each against the suffix beginning at the current cursor.
Each literal is aligned against the suffix beginning at that cursor.
$\operatorname{LCSBlocks}(a,b)$ partitions the pairwise LCS alignment into maximal runs whose matched indices are consecutive in both strings.
These contiguous blocks replace the old literal in alignment order.
An empty alignment leaves the cursor unchanged.
Otherwise, the cursor advances past each block's earliest available occurrence in the new output.

\begin{figure*}[htbp]
    \centering
    \includegraphics[width=0.98\linewidth]{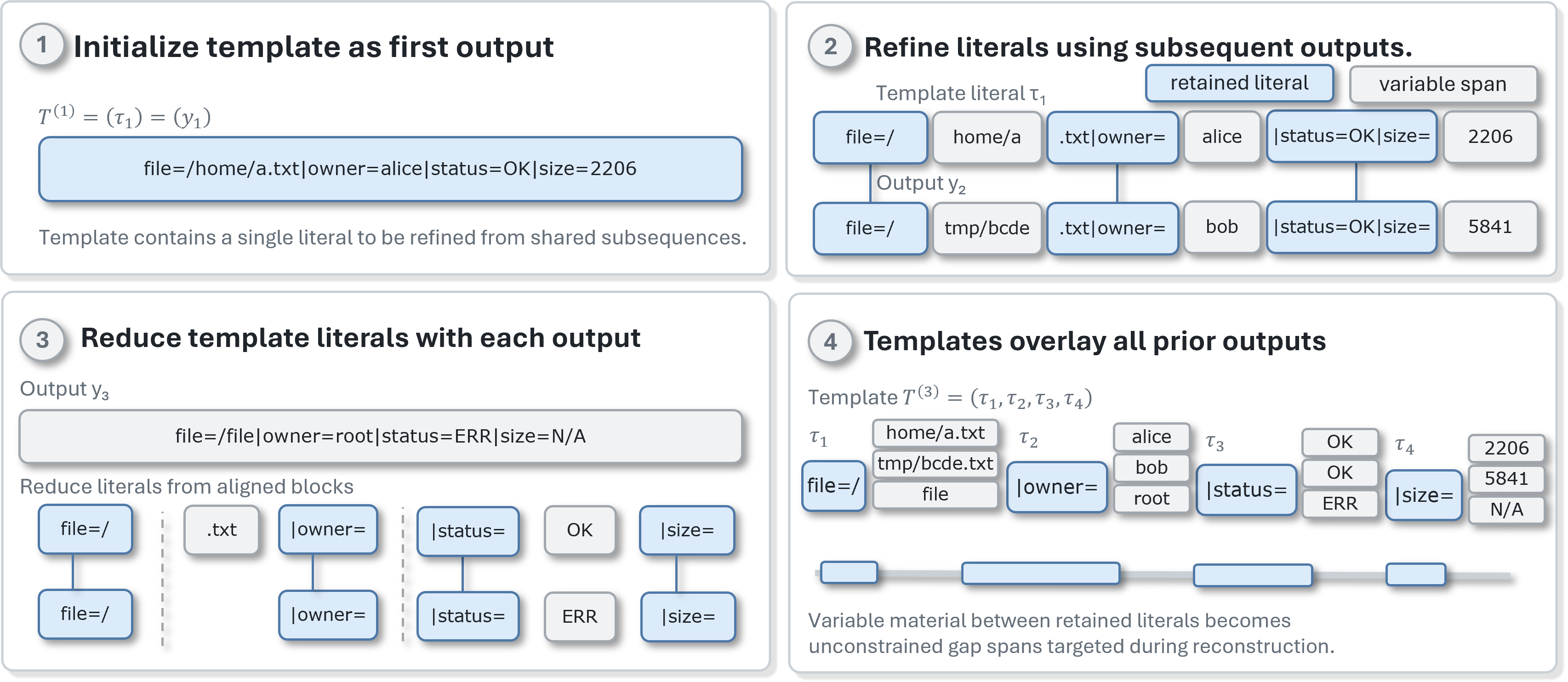}
    \caption{Greedy refinement of three illustrative outputs. Blue blocks are retained literals. Refinement with the third output removes the previously shared \texttt{txt} while retaining the period. The final \texttt{\textbackslash n} denotes one newline character.}
    \label{fig:template-construction-refinement}
\end{figure*}

$\operatorname{FirstOccurrence}(y,\beta,p)$ returns the earliest index $q\geq p$ at which the substring $\beta$ begins in $y$.
\Cref{fig:template-construction-refinement} illustrates this refinement process wherein the first two outputs retain \texttt{.txt}, while the third retains only the shared period at that position.

\noindent\begin{minipage}{\linewidth}
\begin{lemma}
\label{lem:alg1}
The template returned by \Cref{alg:template-extraction} overlays every output in its construction group.
\end{lemma}
\begin{proof}
The claim is vacuous for an empty collection. If its first output is empty, the algorithm returns the empty template, which overlays every output.
Otherwise, $T^{(1)}=(y_1)$ overlays $y_1$.

Assume $T^{(i-1)}=(\tau_1,\ldots,\tau_k)$ overlays all previously processed outputs.
Each old segment $\tau_j$ is replaced by a sequence $B_j$ of ordered, non-overlapping substrings of $\tau_j$, or by the empty sequence.
Replacing an occurrence of $\tau_j$ in any previous output by the corresponding occurrences of its retained substrings preserves their order and non-overlap.
Thus the concatenated blocks $T^{(i)}=B_1\mathbin{\Vert}\cdots\mathbin{\Vert}B_k$ still overlays every previously processed output.

In the new output $y_i$, the LCS alignment supplies ordered, non-overlapping occurrences of each $B_j$ in the suffix available when $\tau_j$ is processed.
Choosing the earliest available occurrence of a block places its endpoint no later than the endpoint supplied by that alignment. Inductively, each subsequent aligned occurrence remains available.
The cursor is carried between old segments, so the blocks also occur in global append order without overlap.
Consequently, $T^{(i)}$ also overlays $y_i$, completing the induction.
\end{proof}
\end{minipage}
\par
{\clubpenalty=10000
With output lengths bounded by $L$, the current literals have total length at most $L$.
Quadratic dynamic programming therefore spends $O(L^2)$ alignment time per refinement, giving $O(nL^2)$ alignment time per group.
\par}

\paragraph{Inventories and assignment.}
For command identifier $c$, the bank $\mathcal T^c=\{T_0,T_1,\ldots,T_{k_c}\}$ contains the non-empty templates constructed from its input groups and the empty template.
For a reference output $y$, the admissible indices are
\[
\mathcal A^c(y)=\{j\in\{0,\ldots,k_c\}:T_j\preceq y\}.
\]
We assign the template with maximum literal coverage:
\begin{equation}
\label{eq:template-assignment}
j^\star=\argmax_{j\in\mathcal A^c(y)}\operatorname{cov}(T_j),
\qquad T^\star=T_{j^\star}.
\end{equation}
Maximum-coverage assignment therefore minimizes the remaining output among admissible inventory templates.
This single assignment supplies both the template used to construct each TGR training target and the corresponding label for input-conditioned template selection.

\subsection{Template Gap Reconstruction}
\label{sec:method_reconstruction}

For a training pair $(x,y)$, we use $T=T^\star$ from \Cref{eq:template-assignment}.
When repeated literals permit multiple decompositions, the earliest-match rule in \Cref{sec:method_extraction} defines consistent gaps $s_0,\ldots,s_k$ from the unmatched prefix, intervening text, and suffix.
Each generated span must retain its insertion position, including when the span is empty.
We associate each gap with a sentinel and represent the generation target as
\begin{equation}
\label{eq:gap-target}
\widetilde T
=\mathrm{[S0]}s_0\mathrm{[S1]}s_1
\cdots\mathrm{[Sk]}s_k.
\end{equation}
The final sentinel begins the trailing gap $s_k$, after which ordinary response termination ends generation.

The reconstruction model receives the input $x$ and template $T$, with the same sentinels inserted at the corresponding gap positions in the template.
This identifies each span's position relative to the surrounding literals, while $x$ supplies the command's options and resource arguments.
TGR aims to maximize the conditional likelihood of the gap target under the reconstruction model $\theta$:
\begin{equation}
\label{eq:tgr-likelihood}
P_\theta(\widetilde T\mid x,T)
=\prod_{t=1}^{m}
P_\theta(\widetilde T_t\mid x,T,\widetilde T_{<t}),
\end{equation}
where $m$ is the tokenized target length, $\widetilde T_t$ is its $t$-th token, and $\widetilde T_{<t}$ contains the preceding reference target tokens.

During generation, the sentinels delimit the predicted gaps $\widehat s_0,\ldots,\widehat s_k$.
For a complete gap sequence, reconstruction interleaves these predictions with the supplied template literals:
\begin{equation}
\label{eq:output-reconstruction}
\widehat y
=\widehat s_0\tau_1\widehat s_1\tau_2
\cdots\tau_k\widehat s_k.
\end{equation}
The literals are copied rather than predicted, leaving their intervening text to the model.
Substituting the reference gaps recovers $y$ by \Cref{eq:template-decomposition}.

\paragraph{Worked example.}
Consider the shell input
\[
x=\texttt{stat --printf="size=\%s\textbackslash n" ./a.txt}
\]
with recorded output $y=\texttt{size=128\textbackslash n}$, where \texttt{\textbackslash n} denotes one newline character.
For $T=(\texttt{size=},\texttt{\textbackslash n})$, the reference gaps are $s_0=s_2=\epsilon$ and $s_1=\texttt{128}$.
The template is supplied as
\[
\texttt{[S0]size=[S1]\textbackslash n[S2]},
\]
and the generation target is
\[
\widetilde T=\texttt{[S0][S1]128[S2]}.
\]
The adjacent first two sentinels preserve the empty leading gap, while the final sentinel begins the empty trailing gap.
Generating \texttt{128} for the middle gap and leaving the boundary gaps empty reconstructs the recorded output by copying \texttt{size=} and the newline around the predicted value.

\begin{figure*}[htbp]
    \centering
    \includegraphics[width=0.98\linewidth]
        {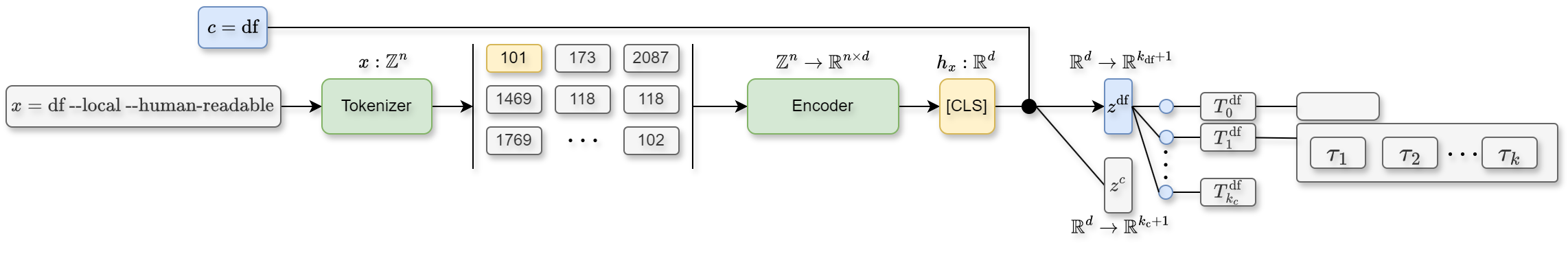}
    \caption{Input-conditioned template selection.
    A shared encoder represents the submitted input, and a command-specific head scores its template inventory, including the empty-template FOD class.}
    \label{fig:5_template_selection}
\end{figure*}

\subsection{Template Selection}
\label{sec:method_selection}
The assignment in \Cref{eq:template-assignment} depends on the reference output, which is unavailable when choosing a template for a submitted input.
We therefore train a selector to predict the assigned index $j^\star$ from $x$.
Using this assignment aligns selector supervision with the maximum-coverage admissible template used to construct each reconstruction-training target.

The command identifier $c$ determines the candidate inventory $\mathcal T^c$, while the submitted options and arguments can distinguish response forms within that inventory.
We therefore encode the complete input $x$ with a Transformer encoder, obtaining a pooled representation $h_x\in\mathbb R^d$.
The encoder is shared across commands, and a command-specific linear head scores the entries in $\mathcal T^c$.
Sharing the encoder allows input features to be learned across commands, while the separate heads restrict each decision to the corresponding template classes.
\Cref{fig:5_template_selection} illustrates this architecture.

Template selection aims to maximize the conditional likelihood of the assigned index under the selector $\phi$:
\begin{equation}
\label{eq:selector-likelihood}
P_\phi(j^\star\mid x,c)
=\operatorname{softmax}(W_ch_x+b_c)_{j^\star},
\end{equation}
where $W_c\in\mathbb R^{(k_c+1)\times d}$ and $b_c\in\mathbb R^{k_c+1}$ parameterize the command-specific head.
The selector parameters $\phi$ comprise the shared encoder and these heads and are distinct from the reconstruction parameters $\theta$.
When no non-empty template overlays the reference, the assigned index is $j^\star=0$.
Including the empty-template class therefore provides supervision for full-output fallback within the same classification task, rather than requiring a non-empty template for every input.

At inference, the selector chooses the template with the highest conditional probability:
\begin{equation}
\label{eq:inference-selection}
\widehat j
=\argmax_{0\leq j\leq k_c}P_\phi(j\mid x,c),
\qquad
\widehat T=T_{\widehat j},
\end{equation}
with ties resolved by the fixed inventory order.

For $\widehat j>0$, the reconstruction model receives $x$ and $\widehat T$, serialized with the same sentinel convention used during training.
The predicted template replaces the reference assignment; gap generation and assembly follow \Cref{eq:output-reconstruction}.
For $\widehat j=0$, the same jointly trained reconstruction checkpoint instead receives $x$ alone and directly generates the complete response through its ordinary FOD interface.
This retains full-output prediction for inputs routed to the empty-template class without requiring a separate reconstruction model.

%% file: content/4_eval_v2.tex
\section{Evaluation}
\label{sec:5_eval}
We evaluate whether reusable output structure improves response fidelity at fixed reconstruction weights and where that benefit is concentrated.
Complete-response and gap-only analyses separate assembled-response quality from the gap prediction needed for TGR across template coverage and gap length.
Tokenized reference-target lengths quantify the text assigned to generation.

\subsection{Experimental Setup}
\label{sec:eval-setup}

\paragraph{Data and training.}
We construct the template bank from 18.7 million shell outputs gathered across execution contexts.
The configurations in \Cref{tab:microvm_configs} span 25 filesystem images with different users, environment variables, and contents.
The construction procedure groups canonicalized inputs by the observed exit code and requires at least 32 execution records per group.
The resulting bank is fixed before model training and evaluation.
A separate one-million-record corpus from a single target environment supplies reconstruction and selection training examples.
That environment and its synthesized executions are excluded from bank construction, although individual features may coincide across environments.

\begin{table}[tb]
    \centering
    \caption{MicroVM configuration features for each execution environment across 25 filesystems.}
    \label{tab:microvm_configs}
    \begin{tabular}{ccccc}
    \toprule
         User & N CPUs & Kernel & Memory & Disk \\
    \midrule
         root       & 1 & 5.15.220 & 0.5G & 0.5G \\ 
         non-root   & 2 & 6.1.128  & 1G   & 1G \\
                    &   & 6.8.0    & 2G   & 2G \\
    \bottomrule
    \end{tabular}
\end{table}

We evaluate Qwen3-0.6B \cite{yang2025qwen3}, ByT5-small and ByT5-base \cite{xue2022byt5}, and CodeT5-small and CodeT5-base \cite{wang2021codet5}, covering decoder-only and encoder-decoder architectures with subword or byte-level representations.
Each backbone uses a single joint checkpoint trained on 62,500 minibatches of 32 samples, with a balanced mixture of FOD and TGR examples.
The selector uses a shared CodeBERT encoder \cite{feng2020codebert} and 86 command-specific heads.
Training uses Adam, an initial learning rate of $10^{-4}$, a 0.05 warm-up fraction, and linear decay.
TGR targets use maximum-coverage admissible assignment from \Cref{eq:template-assignment}.
GPT-4o-mini receives an environment-description prompt, while Cowrie mirrors the target's visible kernel, user, environment variables, and filesystem \cite{cowrie}.
Both provide contextual baselines.
\Cref{sec:appendix} documents the baseline prompt and filesystem.

\paragraph{Output generation methods.}
FOD generates the complete response from the submitted input.
Selected TGR uses an input-conditioned selector in which a nonempty selection supplies the corresponding template; selecting $T_0$ invokes FOD.
Oracle TGR instead uses the reference to select the maximum-coverage admissible template, retaining the same generation and fallback procedures; this gives an upper bound on TGR performance.
All three methods use the same joint checkpoint for each backbone and hold the reconstruction weights fixed.
Consequently, the comparison evaluates alternative inference interfaces available to that checkpoint.

\paragraph{Metrics.}
Each output-generation method is evaluated on 1,000 test inputs using the outputs produced in the target environment.
Output edit similarity (O-ES) is computed per record and then averaged across records; by definition, the O-ES score equals $1$ when both strings are empty.
\begin{equation}
\label{eq:eval-output-edit-similarity}
\operatorname{O\text{-}ES}(y,\widehat y)
=1-\frac{d_{\mathrm{Lev}}(y,\widehat y)}
{\max(|y|,|\widehat y|,1)}.
\end{equation}
This edit-similarity measure provides a direct quantitative comparison to the output observed from the targeted system. In addition, we report n-gram-based metrics to assess sequential fidelity. Mean BLEU-4 is computed using case-sensitive 13a tokenization, exponential smoothing, and effective n-gram orders up to four \cite{papineni2002bleu,chen2014smoothing}. All BLEU scores are obtained via this record-level computation.
ROUGE-L F1 \cite{lin2004rouge} is calculated over the same token sequence and then averaged per record. For both token-based metrics, the score is set to 1 when comparing two empty sequences and to 0 when exactly one sequence is empty.
O-ES is reported on a 0--1 scale, whereas BLEU-4 and ROUGE-L F1 are reported on a 0--100 scale, with larger values indicating better performance.
For each metric, the overall comparison aggregates results over all 1,000 inputs.
Reported means across models are obtained by averaging five independently computed scores.

\paragraph{Template selection.}
The CodeBERT selector obtains a template hit rate (THR) of 94.2\% and a false template rate (FTR) of 4.0\%.
THR credits an overlaying nonempty template, or $T_0$ when no nonempty template is admissible.
FTR counts incompatible nonempty selections over all inputs, identifying copied literals that already conflict with the reference.
A compatible selection need not be the maximum-coverage template, and the predicted gaps still determine the assembled response.
THR therefore measures compatibility rather than exact agreement with the training label or the fraction of correct final responses.
The selected--oracle comparison measures the consequence of changing the assignment policy.

\subsection{Full Output Evaluation}
\label{sec:eval-reconstruction}

\begin{figure*}[t]
    \centering
    \begin{subfigure}[t]{0.49\linewidth}
        \centering
        \includegraphics[width=\linewidth]{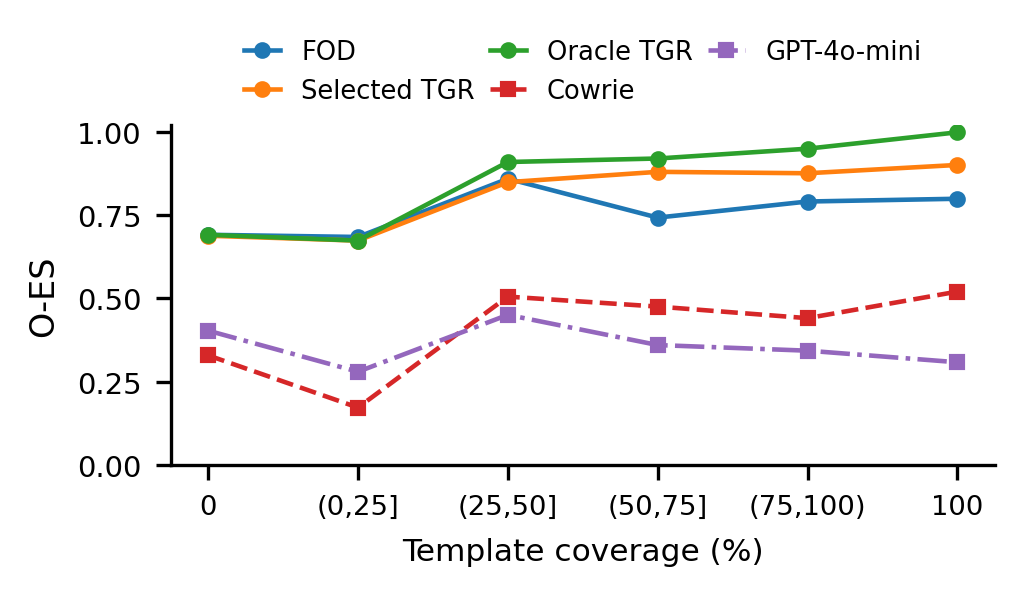}
        \caption{Output edit similarity.}
    \end{subfigure}\hfill
    \begin{subfigure}[t]{0.49\linewidth}
        \centering
        \includegraphics[width=\linewidth]{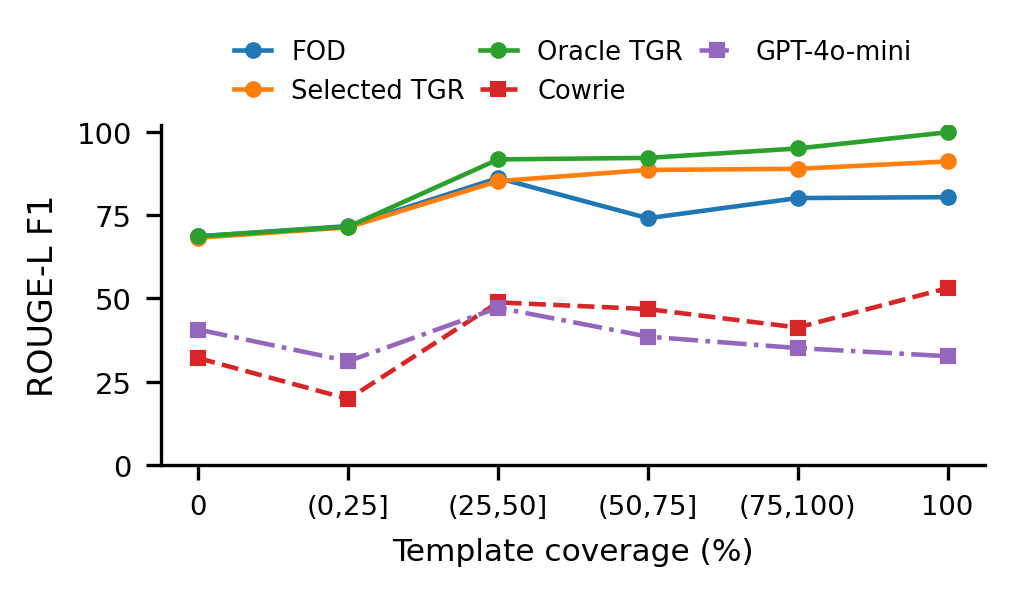}
        \caption{ROUGE-L F1.}
    \end{subfigure}
    \caption{Mean complete-response fidelity by reference-informed template coverage.    Trained curves average five models fine-tuned for FOD/TGR.
    }
    \label{fig:coverage-fidelity}
    \label{fis:es_line}
\end{figure*}

\paragraph{Overall means.}
\Cref{tab:editsim_all} reports mean O-ES of 0.789, 0.808, and 0.824 for FOD, selected TGR, and oracle TGR.
The corresponding mean BLEU-4 scores are 76.43, 78.21, and 79.93, and ROUGE-L scores are 79.67, 81.82, and 83.31.
Selected TGR thus improves the mean by 0.019 O-ES, 1.78 BLEU-4 points, and 2.15 ROUGE-L points over FOD.
The agreement across character and token metrics supports a benefit to the complete response under multiple notions of textual similarity.
Oracle assignment adds an additional 0.016 O-ES, 1.72 BLEU-4 points, and 1.49 ROUGE-L points over the selected TGR, indicating that the choice of supplied structure remains consequential.

All five models show improvement, with varying magnitudes (\Cref{tab:all-input-character}).
Selected O-ES gains range from 0.012 for CodeT5-base to 0.036 for Qwen3-0.6B.
Qwen3-0.6B also has the largest BLEU-4 and ROUGE-L gains, at 2.98 and 3.71 points, while CodeT5-base retains the highest absolute scores under all three generation conditions.
Thus, the model that benefits most from the interface is not necessarily the one with the strongest final response fidelity.
These within-checkpoint improvements support the usefulness of supplied structure across the evaluated architectures without implying a uniform gain or a model-size scaling relation.

Cowrie and GPT obtain O-ES of 0.354 and 0.488, respectively, on the same 1,000 inputs, well below the trained-model means under all three metrics.
Because these baselines differ in training and conditioning, the within-checkpoint comparisons provide more direct evidence for the reconstruction interface and are provided to contextualize the trained model performance.

\begin{table}[tb]
    \centering
    \small
    \setlength{\tabcolsep}{3pt}
    \caption{Mean complete-response fidelity, averaged across five models fine-tuned for FOD/TGR.
    }
    \label{tab:editsim_all}
    \begin{tabular}{@{}lrrr@{}}
    \toprule
    Method & O-ES & BLEU-4 & ROUGE-L F1 \\
    \midrule
    Cowrie & 0.354 & 29.12 & 35.52 \\
    GPT-4o-mini & 0.488 & 42.63 & 50.07 \\
    \midrule
    FOD & 0.789 & 76.43 & 79.67 \\
    Selected TGR & 0.808 & 78.21 & 81.82 \\
    Oracle TGR & \textbf{0.824} & \textbf{79.93} & \textbf{83.31} \\
    \bottomrule
    \end{tabular}
\end{table}

\paragraph{By template coverage.}
For a nonempty reference, available template coverage is given as
\begin{equation}
\label{eq:eval-available-coverage}
\rho(y)=100\,\frac{\operatorname{cov}(T^\star)}{|y|},
\end{equation}
where $T^*$ is the extracted template with the highest coverage for that output.
Every method is grouped by this same reference-informed quantity, including FOD and the contextual baselines.
Coverage therefore describes the structure available for that output rather than that which was utilized.
\Cref{fig:coverage-fidelity} reports accuracy for records bucketed by their template coverage.

The most pronounced advantages of the selected-TGR approach are observed at coverage levels exceeding 50\%.
In particular, selected O-ES attains mean values of 0.879 and 0.875 in the $(50,75]\%$ and $(75,100)\%$ strata, respectively, compared with 0.742 and 0.790 for FOD.
The resulting absolute improvements (0.137 and 0.085) are greater than the overall mean gain, and both token-level metrics likewise increase within these strata (\Cref{tab:coverage-all}). Collectively, these strata indicate substantial performance benefits at higher coverage levels.

The lower-coverage groups distinguish two limitations.
At $(25,50]\%$, oracle O-ES improves over FOD from 0.859 to 0.909, but selected TGR scores 0.848.
An admissible reference assignment is useful on average in this group, yet learned selection does not realize that benefit.
At $(0,25]\%$, both selected and oracle TGR score 0.673, slightly below FOD's 0.684.
Here, supplying the maximum-coverage admissible template is itself insufficient to improve the group means, as correcting incompatible selections alone would not explain or remove the regression.

At the endpoints, oracle TGR and FOD agree at 0.691 O-ES when no nonempty template is available, consistent with their shared $T_0$ fallback.
The fully templated group reaches 0.998 under oracle assignment and 0.900 under learned selection, versus 0.799 for FOD.
Because its reference gaps contain no characters, this group primarily tests selection and reconstruction of supplied structure, including completion of the sentinel interface.

\paragraph{By reference-gap length.}
Because coverage can conceal a large absolute prediction problem in a long response, we also group complete-response scores by total reference-gap length,
\begin{equation}
\label{eq:eval-reference-gap-length}
\ell_{\mathrm{gap}}(y)=\sum_{i=0}^{k}|s_i|
=|y|-\operatorname{cov}(T^\star).
\end{equation}
This is the number of reference characters outside the assigned literals, summed across all gap slots.

\Cref{tab:mean-response-gap-length} shows that TGR's advantage over FOD is concentrated on gaps whose total length is $< 64$.
For the shorter-gap buckets, selected and oracle TGR achieve mean O-ES of 0.903 and 0.964, respectively, compared with 0.812 for FOD.
Selected TGR also improves BLEU-4 and ROUGE-L by 8.88 and 9.25 points, respectively.
For the longer-gap buckets, the methods perform more similarly: mean O-ES is 0.804 for selected TGR, 0.822 for oracle TGR, and 0.809 for FOD, while selected-TGR differences narrow to $-0.79$ BLEU-4 points and $+1.05$ ROUGE-L points (\Cref{tab:residual-length-all}).
These results indicate that reusable output structure provides substantial gains when relatively little gap text remains to be generated, while TGR and FOD achieve comparable fidelity for longer gap sequences.

\begin{table}[tb]
    \centering
    \small
    \setlength{\tabcolsep}{3pt}
    \caption{Mean full output similarity by reference-gap length, averaged across five models fine-tuned for FOD/TGR.
    }
    \label{tab:mean-response-gap-length}
    \begin{tabular}{@{}lrrrrr@{}}
    \toprule
    & \multicolumn{5}{c}{Gap length (characters)} \\
    \cmidrule(lr){2-6}
    Condition & 0 & 1--16 & 17--64 & 65--256 & 257+ \\
    \midrule
    \multicolumn{6}{@{}l}{\textit{O-ES}} \\
    FOD & 0.799 & 0.855 & 0.783 & 0.827 & \textbf{0.790} \\
    Selected TGR & 0.900 & 0.938 & 0.870 & 0.843 & 0.765 \\
    Oracle TGR & \textbf{0.998} & \textbf{0.958} & \textbf{0.935} & \textbf{0.872} & 0.772 \\
    \midrule
    \multicolumn{6}{@{}l}{\textit{BLEU-4}} \\
    FOD & 77.63 & 82.77 & 75.39 & 77.53 & \textbf{76.98} \\
    Selected TGR & 89.25 & 89.44 & 83.74 & 79.50 & 73.43 \\
    Oracle TGR & \textbf{99.73} & \textbf{92.81} & \textbf{91.24} & \textbf{81.69} & 74.43 \\
    \midrule
    \multicolumn{6}{@{}l}{\textit{ROUGE-L F1}} \\
    FOD & 80.35 & 86.16 & 78.27 & 84.02 & \textbf{80.69} \\
    Selected TGR & 91.07 & 93.60 & 87.87 & 87.07 & 79.74 \\
    Oracle TGR & \textbf{99.81} & \textbf{95.87} & \textbf{93.96} & \textbf{88.58} & 80.64 \\
    \bottomrule
    \end{tabular}

\end{table}

\subsection{Gap Prediction Evaluation}
\label{sec:eval-gap-fidelity}
We score generated gaps without credit for supplied literals, evaluating text outside the maximum-coverage admissible template.
Oracle TGR fixes the same reference-informed template decomposition for all five models, separating gap prediction from learned template selection.
For corresponding gaps $s_i$ and $\widehat s_i$,
\begin{equation}
\label{eq:eval-gap-edit-similarity}
\operatorname{Gap\text{-}ES}
=1-\frac{\sum_i d_{\mathrm{Lev}}(s_i,\widehat s_i)}
{\max(1,\sum_i\max(|s_i|,|\widehat s_i|))}.
\end{equation}
All expected gaps remain in the reference, including missing trailing predictions.
Gap BLEU-4 and ROUGE-L use case-sensitive 13a tokens and within-slot counts, preventing matches across gap boundaries (\Cref{sec:append-gap-protocol}).



\paragraph{By template coverage.}
\Cref{tab:gap-coverage} uses the four partial-coverage groups shared across models.
The $(25,50]\%$ group has the highest mean under all three gap metrics.
Mean Gap ES increases from 0.472 in $(0,25]\%$ to 0.865 in $(25,50]\%$, then decreases to 0.828 and 0.797 in the higher-coverage groups.
Greater available coverage is therefore not accompanied by uniformly higher gap fidelity across these inputs.
In particular, oracle complete-response O-ES rises from 0.909 in $(25,50]\%$ to 0.949 in $(75,100)\%$ while Gap ES falls from 0.865 to 0.797.
The assembled response can benefit from more supplied correct text even when the remaining text is harder to predict accurately.
The 0\% group reduces to FOD and the 100\% group has no nonempty reference gap, resulting in there being no difference from the full output evaluation in \Cref{tab:mean-response-gap-length}.

\begin{table}[tb]
    \centering
    \small
    \setlength{\tabcolsep}{4pt}
    \caption{Mean oracle-TGR gap fidelity by reference-informed template coverage, averaged across five models fine-tuned for FOD/TGR.
    }
    \label{tab:gap-coverage}
    \label{tab:gap-metric-coverage}
    \begin{tabular}{@{}lrrr@{}}
    \toprule
    Coverage (\%) & Gap ES & BLEU-4 & ROUGE-L F1 \\
    \midrule
    $(0,25]$ & 0.472 & 44.98 & 52.02 \\
    $(25,50]$ & 0.865 & 83.97 & 87.16 \\
    $(50,75]$ & 0.828 & 73.29 & 80.32 \\
    $(75,100)$ & 0.797 & 73.90 & 78.92 \\
    \bottomrule
    \end{tabular}
\end{table}

\paragraph{By reference-gap length.}
\Cref{tab:mean-gap-length} instead groups gap-only scores by total reference-gap characters.
Fully templated responses (those with zero gap characters) are absent because they require no nonempty gap prediction.
Mean Gap ES decreases from 0.860 for 1--16 characters to 0.510 for 257--1,024, and gap ROUGE-L decreases from 85.34 to 55.07.
Mean BLEU-4 likewise decreases from 83.31 to 50.61 across these groups.
Unlike the coverage stratification, all three backbone-mean gap metrics decline at every reported length boundary.
The character-level decline is also present for each individual backbone (\Cref{tab:gap-length-all}), although individual token metrics have exceptions.
Gap length thus provides a more consistent descriptive ordering of gap difficulty than coverage.

\begin{table}[tb]
    \centering
    \small
    \setlength{\tabcolsep}{4pt}
    \caption{Mean oracle-TGR gap fidelity by reference-gap length, averaged across five models fine-tuned for FOD/TGR.
    }

    \label{tab:mean-gap-length}
    \begin{tabular}{@{}lrrr@{}}
    \toprule
    Gap characters & Gap ES & BLEU-4 & ROUGE-L F1 \\
    \midrule
    1--16 & 0.860 & 83.31 & 85.34 \\
    17--64 & 0.774 & 74.89 & 77.69 \\
    65--256 & 0.699 & 60.14 & 70.33 \\
    257+ & 0.510 & 50.61 & 55.07 \\
    \bottomrule
    \end{tabular}
\end{table}

\subsection{Target Generation Length}
\label{sec:eval-generation-length}
The one-million-record training corpus supplies tokenized FOD references and their oracle TGR gap targets.
We compare each tokenizer against its own FOD references, since byte-level and subword positions represent different amounts of output text.
\Cref{tab:reconstruction_efficiency} reports 26.3--38.3\% shorter reference targets, with the largest reduction for byte-level ByT5.
The absolute reductions are 11.44 positions for Qwen3, 49.82 for ByT5, and 12.67 for CodeT5 per record on average.
That all three remain positive after counting sentinels shows that delimiter overhead does not erase the mean reduction in this corpus.
The percentage difference across tokenizer families also shows why a character-coverage objective cannot be treated as a tokenizer-independent estimate of generation cost.
These measurements concern reference targets under reference-informed assignment where shorter targets provide a concrete efficiency opportunity for small shell models.

\begin{table}[htbp]
    \centering
    \small
    \caption{Mean lengths of the gap sequence for the optimal template.
    TGR counts include sentinel delimiters.
    Reduction is $100(1-\mathrm{TGR}/\mathrm{FOD})$.}
    \label{tab:reconstruction_efficiency}
    \begin{tabular}{@{}lrrr@{}}
    \toprule
    Tokenizer & FOD & TGR & Reduction \% \\
    \midrule
    Qwen3 & 42.89 & 31.45 & 26.7 \\
    ByT5 & 129.98 & 80.16 & 38.3 \\
    CodeT5 & 48.16 & 35.49 & 26.3 \\
    \bottomrule
    \end{tabular}
\end{table}

%% file: content/5_discussion.tex
\section{Discussion}
\label{sec:discussion}
The results support output representation as a design choice for small language models used in shell simulation.
Selected TGR raises mean O-ES from 0.789 to 0.808 and improves all three aggregate metrics for every evaluated backbone (\Cref{tab:editsim_all,tab:all-input-character}).
With reconstruction weights fixed, these gains reflect changes to the information provided and the text generated within the jointly trained model.

\paragraph{Representation and reconstruction capacity.}
ByT5-small with selected TGR exceeds ByT5-base with FOD under all three overall metrics, including O-ES of 0.794 versus 0.784.
This within-family finding indicates that the choice of output representation can mitigate the empirically observed similarity gap across reconstruction model sizes, although it does not establish a scaling law nor constitute a general replacement for increased parameterization. The selector module introduces additional parameters and inference-time overhead.
Deployment decisions should evaluate the joint configuration of the reconstruction model, selection mechanism, and output representation with respect to output similarity and available computational and memory budgets.

Complete-response scores credit copied template literals, while gap scores evaluate predictions without bias from template selection.
Mean oracle Gap ES falls from 0.865 in the $(25,50]\%$ coverage group to 0.797 in $(75,100)\%$, even as complete-response O-ES increases across those groups.
A better assembled response can therefore coexist with less accurate gap prediction.
Mean gap similarity also declines across reference-gap-length groups, identifying gap generation as a source of error.

Template compatibility permits error-free copying but does not identify the best reconstruction option.
These conditions are asymmetrical in impact: an unnecessary $T_0$ choice forgoes possible reuse, while an incompatible nonempty template commits the response to literals that gap prediction cannot revise.
Minimizing the false template rate is thus imperative to bridge the gap between selected and oracle TGR.
Selection based on predicted reconstruction quality for a given backbone could keep FOD as a candidate even when a nonempty template is admissible.

\paragraph{Target length and deployment cost.}
The 26.3--38.3\% shorter mean reference targets include sentinels and quantify fewer generation positions on the training corpus.
Maximum character coverage need not minimize tokenized length because segmentation introduces delimiters and tokenizers encode literals and gaps differently.
Actual generated lengths can differ from reference lengths, while learned selection changes the templates used and the selector and template context add computation.

The bank excludes the target environment, but both learned components receive target-environment training.
As such, these results concern a specialized simulator.
Similarly, this study examines strategies for enhancing output generation under single-input conditions and does not consider modifications to the execution environment that could influence subsequent interactions in multi-input sessions.
Nevertheless, the observed performance gains complement approaches that condition upon and update the execution environment and offer a mechanism that can be integrated alongside such methods for output generation.
MANTIS evaluates consistency across command sequences using explicit filesystem state and interaction-history management \cite{wei2026mantis}.
Session-level evaluation with explicit state conditioning and checks of consequential fields would connect response similarity to stateful honeypot behavior.

%% file: content/6_conclusion.tex
\section{Conclusion}
\label{sec:conclusion}
We introduced template gap reconstruction (TGR) to improve output fidelity and reduce the text predicted by small language models for execution-free shell simulation.
Templates extracted from 18.7 million executions across heterogeneous environments supply recurring literal spans, while the model generates the gaps needed to complete each response.
An input-only selector chooses the template, with the empty template preserving full-output decoding (FOD) through the same jointly trained reconstruction checkpoint.

For each of five backbones spanning approximately 60--600 million parameters, selected TGR improves mean complete-response fidelity over FOD at fixed reconstruction weights.
The benefits are concentrated in responses with substantial reusable structure and short gaps.
Oracle TGR improves O-ES in every reported coverage group above 25\%, while learned selection yields consistent gains above 50\%.
For responses with 0--64 reference-gap characters, oracle TGR achieves mean O-ES of 0.964 compared with 0.812 for FOD, with closer performance for longer gaps.

TGR also shortens mean tokenized reference targets by 26.3--38.3\% across tokenizer families, including sentinel overhead.
Together, these findings support reusable output structure as a practical design choice for compact shell simulators: supplied literals improve response fidelity while reducing the text assigned to autoregressive generation.
Template selection and gap prediction remain central to the quality of the assembled response.

%% file: content/7_appendix.tex
\section{Supplementary Material}
\label{sec:appendix}
\begin{figure*}[htb]
    \centering
    \includegraphics[width=0.95\linewidth]{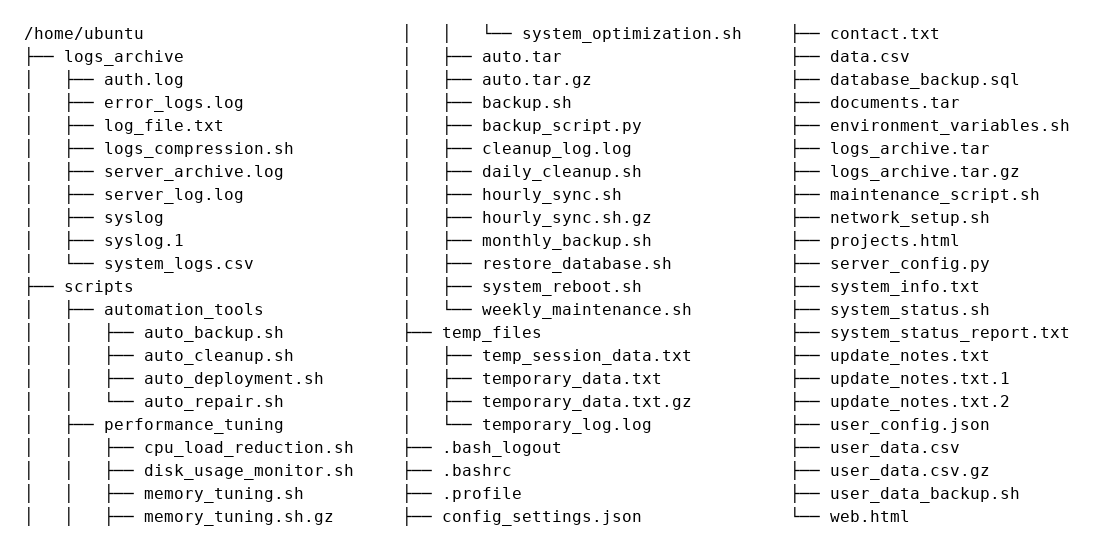}
    \caption{Working directory of execution environment}
    \label{fig:fs1}
\end{figure*}

The contextual baselines and trained reconstruction models have different sources of information about the target environment.
GPT-4o-mini receives an explicit environment description and the submitted input, while the reconstruction backbones are specialized using recorded executions from that environment (\Cref{sec:eval-setup}).
\Cref{fig:append_llm} documents the prompted interface, and \Cref{fig:fs1} shows the target working directory.
These details help interpret the baseline comparison in that matching visible environment information does not make the training or conditioning procedures equivalent.

\begin{figure*}[htb]
    \centering
    \begin{minipage}{\linewidth}
\begin{verbatim}
You are emulating a Linux command-line interface for a user to 
interact with. Features of this simulated environment are as follows:
{
'arch': 'x86_64',
'kernel': '6.8',
'cpus': '1',
'mem': '216M',
'os_family': 'ubuntu-24',
'cwd': '/home/ubuntu',
...
'env': {'HOME': '/',
        ...
    },
'fs': {'.': {
          'group': 'ubuntu',
          'owner': 'ubuntu',
          'perms': 'drwxrwxr-x',
          'size': 4096,
          'time': '19:43',
          ...},
          ...
    },
 ...
 }
Respond with the output containing exactly the text that would 
be printed to stdout or stderr.
\end{verbatim}
  \end{minipage}
    \caption{Instruction template provided for GPT-4o-mini baseline}
    \label{fig:append_llm}
\end{figure*}

%% file: content/73_append.tex
\section{Scoring and Evaluation Populations}
\label{sec:append-scoring}
The appendices separate three questions: how faithfully the complete response is reproduced, where that fidelity changes across response structures, and how accurately the model predicts the residual text.
Complete-response scores evaluate the assembled output, including copied literals.
Gap-only scores evaluate corresponding generated spans with their boundaries retained, so correct literals cannot compensate for residual errors.
O-ES and Gap ES are character-level scores on a 0--1 scale.
BLEU-4 and ROUGE-L F1 use case-sensitive 13a tokens and are displayed on a 0--100 scale.
All three measure textual agreement with the recorded response, not semantic equivalence or consistency of shell state.

We report the mean of smoothed record-level BLEU-4 scores between the case-sensitive WMT 13a tokenized prediction and reference \cite{papineni2002bleu,chen2014smoothing}.
For each system, the overall and stratified BLEU results average these record scores, and the model means average the separately computed model means.
A pair with no matching unigrams scores zero.
Orders without candidate n-grams are omitted, allowing short matching responses to receive credit without a nonexistent fourth-order denominator.
We explicitly extend the score to one when both token sequences are empty and zero when only one is empty.
The reference is never cropped, and the brevity penalty is applied within every nonempty record.
These mean sentence-level scores use exponential smoothing and effective order as documented for sentence BLEU \cite{chen2014smoothing}.
ROUGE-L uses the same token sequence and averages record-level token-LCS F1.
Two empty token sequences score one, and a pair with only one empty sequence scores zero.
The same empty-agreement convention applies to BLEU-4.

\subsection{Matched Populations and Aggregation}
The overall complete-response comparison includes the same 1,000 inputs for all metrics and systems, including empty references.
Available coverage and reference-gap length are determined by the reference-informed assignment $T^\star$.
Thus, a coverage group describes the structure available for that input even for FOD and the contextual baselines, which do not receive a template.
Reference-gap length is the total number of residual characters, $\sum_i|s_i|$, across all reference gaps.
In the following tables, bold values identify the highest displayed trained-system score in each metric column, with a separate comparison among model-mean conditions where multiple conditions are reported.

\clearpage
\section{Complete-Response Fidelity}
\label{sec:append-complete-results}
\subsection{Overall Scores}
\Cref{tab:all-input-character} shows that the aggregate ordering is shared by every model: selected TGR improves on FOD, and reference-informed oracle TGR improves on selected TGR under all three metrics.
Selected-minus-FOD O-ES gains range from 0.012 for CodeT5-base to 0.036 for Qwen3-0.6B.
The largest gain therefore occurs for a model that is not the best absolute performer.
CodeT5-base retains the highest complete-response score in each generation condition.
This distinguishes benefit from the interface from the ranking of the underlying reconstruction models.

The selected-to-oracle O-ES differences are much more similar across models, ranging from 0.015 to 0.017.
These gaps describe the cost of using the learned selection policy relative to the maximum-coverage reference assignment, including differences between compatible templates and fallback choices.
They cannot be attributed solely to incompatible selections.
Both contextual baselines score below every trained model even under FOD, so their separation from TGR reflects differences in training and conditioning as well as the reconstruction interface.

\begin{table}[H]
\centering
\small
\setlength{\tabcolsep}{4pt}
\caption{Complete-response fidelity.}
\label{tab:all-input-character}
\label{tab:all-response-results}
\label{tab:all-input-results}
\begin{tabular}{@{}llrrr@{}}
\toprule
Model & Condition & O-ES & BLEU-4 & ROUGE-L F1 \\
\midrule
Cowrie & Full response & 0.354 & 29.12 & 35.52 \\
\addlinespace[2pt]
GPT-4o-mini & Full response & 0.488 & 42.63 & 50.07 \\
\midrule
\midrule
\addlinespace[2pt]
Qwen3-0.6B & FOD & 0.775 & 75.50 & 78.41 \\
 & Selected TGR & 0.811 & 78.48 & 82.12 \\
 & Oracle TGR & 0.828 & 80.66 & 83.78 \\
 \midrule
\addlinespace[2pt]
ByT5-base & FOD & 0.784 & 75.62 & 79.22 \\
 & Selected TGR & 0.798 & 76.88 & 80.85 \\
 & Oracle TGR & 0.813 & 78.38 & 82.20 \\
 \midrule
\addlinespace[2pt]
ByT5-small & FOD & 0.775 & 74.42 & 78.44 \\
 & Selected TGR & 0.794 & 76.28 & 80.42 \\
 & Oracle TGR & 0.810 & 77.75 & 81.93 \\
 \midrule
\addlinespace[2pt]
CodeT5-base & FOD & 0.823 & 80.33 & 82.71 \\
 & Selected TGR & 0.835 & 81.57 & 84.33 \\
 & Oracle TGR & \textbf{0.851} & \textbf{83.25} & \textbf{85.82} \\
 \hline
\addlinespace[2pt]
CodeT5-small & FOD & 0.791 & 76.26 & 79.57 \\
 & Selected TGR & 0.804 & 77.85 & 81.36 \\
 & Oracle TGR & 0.820 & 79.61 & 82.79 \\
\midrule
\midrule
Mean & FOD & 0.789 & 76.43 & 79.67 \\
 & Selected TGR & 0.808 & 78.21 & 81.82 \\
 & Oracle TGR & \textbf{0.824} & \textbf{79.93} & \textbf{83.31} \\
\bottomrule
\end{tabular}
\end{table}

\clearpage
\subsection{By Template Coverage}
\label{sec:append-response-coverage}
\label{sec:append-coverage}
\Cref{tab:coverage-all} locates the overall improvements within reference-defined coverage groups.
For O-ES, every model improves with selected TGR in the $(50,75]\%$, $(75,100)\%$, and fully templated groups.
The mean selected-minus-FOD gains in those groups are 0.137, 0.085, and 0.101, respectively.
This consistency supports the usefulness of reusable structure when it accounts for a substantial part of the response.
The low-coverage comparisons also expose the limits of that benefit: selected mean O-ES is 0.011 below FOD in each of the $(0,25]\%$ and $(25,50]\%$ groups.

\begin{table}[H]
\centering
\normalsize
\setlength{\tabcolsep}{4pt}
\caption{Complete-response fidelity by available template coverage.
}
\label{tab:coverage-all}
\label{tab:coverage-mean}
\label{tab:coverage-qwen3}
\label{tab:coverage-byt5_base}
\label{tab:coverage-byt5_small}
\label{tab:coverage-codet5_base}
\label{tab:coverage-codet5_small}
\begin{tabular}{@{}llrrrrrr@{}}
\toprule
& & \multicolumn{6}{c}{Template coverage (\%)} \\
\cmidrule(lr){3-8}
Model & Condition & 0 & $(0,25]$ & $(25,50]$ & $(50,75]$ & $(75,100)$ & 100 \\
\midrule
\multicolumn{8}{@{}l}{\textit{O-ES}} \\
Cowrie & Full response & 0.329 & 0.173 & 0.505 & 0.475 & 0.441 & 0.520 \\
\addlinespace[2pt]
GPT-4o-mini & Full response & 0.403 & 0.280 & 0.451 & 0.360 & 0.343 & 0.309 \\
\midrule
\midrule
\addlinespace[2pt]
Qwen3-0.6B & FOD & 0.682 & 0.623 & 0.818 & 0.706 & 0.743 & 0.743 \\
 & Selected TGR & 0.680 & 0.650 & 0.844 & 0.877 & 0.854 & 0.910 \\
 & Oracle TGR & 0.682 & 0.654 & 0.913 & 0.921 & 0.931 & \textbf{1.000} \\
 \midrule
\addlinespace[2pt]
ByT5-base & FOD & 0.728 & 0.665 & 0.850 & 0.802 & 0.830 & 0.820 \\
 & Selected TGR & 0.727 & 0.674 & 0.824 & 0.872 & 0.884 & 0.900 \\
 & Oracle TGR & 0.728 & 0.673 & 0.907 & 0.908 & 0.940 & \textbf{1.000} \\
 \midrule
\addlinespace[2pt]
ByT5-small & FOD & 0.644 & 0.637 & 0.861 & 0.673 & 0.768 & 0.809 \\
 & Selected TGR & 0.641 & 0.591 & 0.861 & 0.879 & 0.882 & 0.911 \\
 & Oracle TGR & 0.644 & 0.597 & 0.884 & 0.906 & \textbf{0.969} & \textbf{1.000} \\
 \midrule
\addlinespace[2pt]
CodeT5-base & FOD & \textbf{0.736} & \textbf{0.774} & 0.883 & 0.763 & 0.831 & 0.798 \\
 & Selected TGR & 0.731 & 0.754 & 0.839 & 0.872 & 0.904 & 0.890 \\
 & Oracle TGR & \textbf{0.736} & 0.756 & 0.919 & 0.928 & 0.945 & 0.990 \\
 \midrule
\addlinespace[2pt]
CodeT5-small & FOD & 0.664 & 0.722 & 0.882 & 0.768 & 0.781 & 0.824 \\
 & Selected TGR & 0.660 & 0.698 & 0.875 & 0.898 & 0.853 & 0.889 \\
 & Oracle TGR & 0.664 & 0.687 & \textbf{0.921} & \textbf{0.934} & 0.959 & \textbf{1.000} \\
 \midrule
 \midrule
\addlinespace[2pt]
Mean & FOD & \textbf{0.691} & \textbf{0.684} & 0.859 & 0.742 & 0.790 & 0.799 \\
 & Selected TGR & 0.688 & 0.673 & 0.848 & 0.879 & 0.875 & 0.900 \\
 & Oracle TGR & \textbf{0.691} & 0.673 & \textbf{0.909} & \textbf{0.919} & \textbf{0.949} & \textbf{0.998} \\
\bottomrule
\end{tabular}
\end{table}

\clearpage

\begin{table}[H]
\centering
\normalsize
\setlength{\tabcolsep}{4pt}
\ContinuedFloat
\caption[]{Complete-response fidelity by available template coverage (continued).}
\begin{tabular}{@{}llrrrrrr@{}}
\toprule
& & \multicolumn{6}{c}{Template coverage (\%)} \\
\cmidrule(lr){3-8}
Model & Condition & 0 & $(0,25]$ & $(25,50]$ & $(50,75]$ & $(75,100)$ & 100 \\
\midrule
\multicolumn{8}{@{}l}{\textit{BLEU-4}} \\
Cowrie & Full response & 24.50 & 10.10 & 34.04 & 36.89 & 28.77 & 41.98 \\
\addlinespace[2pt]
GPT-4o-mini & Full response & 32.61 & 16.08 & 41.13 & 29.11 & 24.67 & 20.94 \\
\midrule
\midrule
\addlinespace[2pt]
Qwen3-0.6B & FOD & 65.87 & 59.23 & 79.97 & 67.13 & 70.48 & 71.48 \\
 & Selected TGR & 65.83 & 60.41 & 79.65 & 80.75 & 80.33 & 90.66 \\
 & Oracle TGR & 65.87 & 61.33 & 88.11 & 87.03 & 90.02 & \textbf{100.00} \\
 \midrule
\addlinespace[2pt]
ByT5-base & FOD & 69.28 & 61.59 & 83.11 & 74.15 & 78.90 & 80.56 \\
 & Selected TGR & 69.28 & 61.90 & 79.66 & 80.50 & 83.58 & 89.87 \\
 & Oracle TGR & 69.28 & 61.20 & 87.63 & 85.85 & 89.71 & \textbf{100.00} \\
 \midrule
\addlinespace[2pt]
ByT5-small & FOD & 59.74 & 59.04 & 84.91 & 60.51 & 71.67 & 79.35 \\
 & Selected TGR & 59.70 & 53.52 & 84.11 & 82.28 & 82.76 & 90.77 \\
 & Oracle TGR & 59.74 & 52.99 & 85.59 & 85.73 & \textbf{92.79} & \textbf{100.00} \\
 \midrule
\addlinespace[2pt]
CodeT5-base & FOD & \textbf{70.96} & \textbf{76.24} & 86.89 & 69.24 & 78.52 & 76.82 \\
 & Selected TGR & 70.76 & 72.65 & 81.95 & 83.17 & 86.15 & 87.88 \\
 & Oracle TGR & \textbf{70.96} & 72.71 & 88.75 & \textbf{89.72} & 91.27 & 98.65 \\
 \midrule
\addlinespace[2pt]
CodeT5-small & FOD & 62.23 & 69.71 & 87.05 & 68.18 & 72.34 & 79.96 \\
 & Selected TGR & 62.07 & 67.45 & 86.15 & 84.24 & 79.58 & 87.07 \\
 & Oracle TGR & 62.23 & 65.88 & \textbf{89.86} & 88.70 & 91.71 & \textbf{100.00} \\
 \midrule
 \midrule
\addlinespace[2pt]
Mean & FOD & \textbf{65.62} & \textbf{65.16} & 84.39 & 67.84 & 74.38 & 77.63 \\
 & Selected TGR & 65.53 & 63.18 & 82.30 & 82.19 & 82.48 & 89.25 \\
 & Oracle TGR & \textbf{65.62} & 62.82 & \textbf{87.99} & \textbf{87.41} & \textbf{91.10} & \textbf{99.73} \\
\bottomrule
\end{tabular}
\end{table}

\clearpage

\begin{table}[H]
\centering
\normalsize
\setlength{\tabcolsep}{4pt}
\ContinuedFloat
\caption[]{Complete-response fidelity by available template coverage (continued).
}
\begin{tabular}{@{}llrrrrrr@{}}
\toprule
& & \multicolumn{6}{c}{Template coverage (\%)} \\
\cmidrule(lr){3-8}
Model & Condition & 0 & $(0,25]$ & $(25,50]$ & $(50,75]$ & $(75,100)$ & 100 \\
\midrule
\multicolumn{8}{@{}l}{\textit{ROUGE-L F1}} \\
Cowrie & Full response & 32.08 & 19.87 & 48.80 & 46.80 & 41.27 & 53.09 \\
\addlinespace[2pt]
GPT-4o-mini & Full response & 40.65 & 31.18 & 47.21 & 38.54 & 35.10 & 32.67 \\
\midrule
\midrule
\addlinespace[2pt]
Qwen3-0.6B & FOD & 68.93 & 65.26 & 83.05 & 70.82 & 74.97 & 75.97 \\
 & Selected TGR & 68.71 & 68.17 & 84.14 & 87.98 & 87.56 & 92.21 \\
 & Oracle TGR & 68.93 & 69.00 & 92.12 & 91.76 & 94.30 & \textbf{100.00} \\
 \midrule
\addlinespace[2pt]
ByT5-base & FOD & 71.36 & 71.25 & 85.13 & 79.44 & 84.08 & 82.78 \\
 & Selected TGR & 71.27 & 72.45 & 82.53 & 87.62 & 89.40 & 91.61 \\
 & Oracle TGR & 71.36 & 72.36 & 90.93 & 91.32 & 94.36 & \textbf{100.00} \\
 \midrule
\addlinespace[2pt]
ByT5-small & FOD & 63.34 & 67.91 & 87.07 & 67.04 & 78.46 & 82.11 \\
 & Selected TGR & 62.95 & 63.88 & 86.62 & 87.80 & 89.24 & 92.24 \\
 & Oracle TGR & 63.34 & 65.04 & 89.35 & 89.87 & 96.48 & \textbf{100.00} \\
 \midrule
\addlinespace[2pt]
CodeT5-base & FOD & \textbf{73.80} & \textbf{79.12} & 87.53 & 76.52 & 84.39 & 78.42 \\
 & Selected TGR & 72.97 & 77.91 & 84.88 & 89.25 & 91.36 & 90.06 \\
 & Oracle TGR & \textbf{73.80} & 78.23 & \textbf{93.16} & \textbf{94.09} & 94.30 & 99.05 \\
 \midrule
\addlinespace[2pt]
CodeT5-small & FOD & 65.67 & 74.83 & 87.59 & 76.40 & 78.53 & 82.48 \\
 & Selected TGR & 65.14 & 74.18 & 87.66 & 89.92 & 86.71 & 89.21 \\
 & Oracle TGR & 65.67 & 72.75 & 92.82 & 93.62 & \textbf{95.33} & \textbf{100.00} \\
 \midrule
 \midrule
\addlinespace[2pt]
Mean & FOD & \textbf{68.62} & \textbf{71.67} & 86.07 & 74.04 & 80.09 & 80.35 \\
 & Selected TGR & 68.21 & 71.32 & 85.17 & 88.52 & 88.85 & 91.07 \\
 & Oracle TGR & \textbf{68.62} & 71.48 & \textbf{91.68} & \textbf{92.13} & \textbf{94.95} & \textbf{99.81} \\
\bottomrule
\end{tabular}
\end{table}

At zero available coverage, oracle TGR and FOD agree for every model, as expected from their shared empty-template fallback.
Selected scores are slightly lower because selection can still route such an input to a nonempty template.
At 100\% coverage, oracle O-ES is 1.000 for four models and 0.990 for CodeT5-base, while selected scores range from 0.889 to 0.911.
These cases require no nonempty reference gaps, so they principally test access to the appropriate literals and correct completion of the reconstruction interface.
Less than perfect oracle scores in this group result from adding text to the beginnding and end gaps that exist from a fully templated output.

The token metrics broadly corroborate the coverage pattern, but they do not induce identical local rankings.
For example, Qwen3-0.6B improves O-ES from 0.818 to 0.844 in the $(25,50]\%$ group while its BLEU-4 decreases from 79.97 to 79.65.
Reporting both character and token metrics retains these distinctions instead of treating any one measure as a complete account of response fidelity.

\clearpage
\subsection{By Reference-Gap Length}
\label{sec:append-residual-length}
\Cref{tab:residual-length-all} groups complete responses by the absolute amount of text outside the reference-informed template.
Selected TGR improves mean O-ES over FOD by 0.083 for 1--16 residual characters and 0.087 for 17--64, but only by 0.016 for 65--256.
In the longest reported group for outputs with gaps greater than 256 characters, FOD exceeds both selected and oracle TGR under all three model-mean metrics.
Thus the favorable aggregate result does not extend to every residual-length regime, even when an admissible template is supplied.

\begin{table}[H]
\centering
\normalsize
\setlength{\tabcolsep}{4pt}
\caption{Complete-response fidelity by reference-gap length.
Each metric uses the same reference-defined bins and fixed scoring protocol.}
\label{tab:residual-length-all}
\label{tab:gap-length-results}
\label{tab:residual-length-qwen3}
\label{tab:residual-length-byt5_base}
\label{tab:residual-length-byt5_small}
\label{tab:residual-length-codet5_base}
\label{tab:residual-length-codet5_small}
\begin{tabular}{@{}llrrrrr@{}}
\toprule
& & \multicolumn{5}{c}{Template gap length} \\
\cmidrule(lr){3-7}
Model & Condition & 0 & 1--16 & 17--64 & 65--256 & 257+ \\
\midrule
\multicolumn{7}{@{}l}{\textit{O-ES}} \\
Cowrie & Full response & 0.520 & 0.693 & 0.403 & 0.194 & 0.149 \\
\addlinespace[2pt]
GPT-4o-mini & Full response & 0.309 & 0.520 & 0.336 & 0.307 & 0.245 \\
\midrule
\midrule
\addlinespace[2pt]
Qwen3-0.6B & FOD & 0.743 & 0.860 & 0.680 & 0.763 & 0.700 \\
 & Selected TGR & 0.910 & 0.927 & 0.846 & 0.836 & 0.686 \\
 & Oracle TGR & \textbf{1.000} & 0.941 & 0.925 & 0.863 & 0.694 \\
 \midrule
\addlinespace[2pt]
ByT5-base & FOD & 0.820 & 0.885 & 0.820 & 0.858 & 0.806 \\
 & Selected TGR & 0.900 & 0.957 & 0.887 & 0.827 & 0.840 \\
 & Oracle TGR & \textbf{1.000} & 0.951 & \textbf{0.954} & 0.855 & 0.837 \\
 \midrule
\addlinespace[2pt]
ByT5-small & FOD & 0.809 & 0.839 & 0.775 & 0.774 & 0.785 \\
 & Selected TGR & 0.911 & 0.936 & 0.873 & 0.851 & 0.702 \\
 & Oracle TGR & \textbf{1.000} & \textbf{0.991} & 0.925 & 0.860 & 0.695 \\
 \midrule
\addlinespace[2pt]
CodeT5-base & FOD & 0.798 & 0.867 & 0.818 & \textbf{0.902} & 0.853 \\
 & Selected TGR & 0.890 & 0.955 & 0.866 & 0.859 & 0.842 \\
 & Oracle TGR & 0.990 & 0.954 & 0.935 & 0.895 & \textbf{0.866} \\
 \midrule
\addlinespace[2pt]
CodeT5-small & FOD & 0.824 & 0.824 & 0.823 & 0.839 & 0.809 \\
 & Selected TGR & 0.889 & 0.914 & 0.879 & 0.843 & 0.757 \\
 & Oracle TGR & \textbf{1.000} & 0.952 & 0.934 & 0.885 & 0.770 \\
 \midrule
 \midrule
\addlinespace[2pt]
Mean & FOD & 0.799 & 0.855 & 0.783 & 0.827 & \textbf{0.790} \\
 & Selected TGR & 0.900 & 0.938 & 0.870 & 0.843 & 0.765 \\
 & Oracle TGR & \textbf{0.998} & \textbf{0.958} & \textbf{0.935} & \textbf{0.872} & 0.772 \\
\bottomrule
\end{tabular}
\end{table}

\clearpage

\begin{table}[H]
\centering
\normalsize
\setlength{\tabcolsep}{4pt}
\ContinuedFloat
\caption[]{Complete-response fidelity by reference-gap length (continued).}
\begin{tabular}{@{}llrrrrr@{}}
\toprule
& & \multicolumn{5}{c}{Template gap length (characters)} \\
\cmidrule(lr){3-7}
Model & Condition & 0 & 1--16 & 17--64 & 65--256 & 257+ \\
\midrule
\multicolumn{7}{@{}l}{\textit{BLEU-4}} \\
Cowrie & Full response & 41.98 & 52.09 & 27.75 & 11.09 & 12.95 \\
\addlinespace[2pt]
GPT-4o-mini & Full response & 20.94 & 45.04 & 23.21 & 21.36 & 16.60 \\
\midrule
\midrule
\addlinespace[2pt]
Qwen3-0.6B & FOD & 71.48 & 84.09 & 65.53 & 72.51 & 67.26 \\
 & Selected TGR & 90.66 & 88.57 & 79.68 & 78.07 & 64.08 \\
 & Oracle TGR & \textbf{100.00} & 91.62 & 90.04 & 80.46 & 66.22 \\
 \midrule
\addlinespace[2pt]
ByT5-base & FOD & 80.56 & 85.87 & 79.25 & 81.78 & 78.97 \\
 & Selected TGR & 89.87 & 91.57 & 85.57 & 78.27 & 80.89 \\
 & Oracle TGR & \textbf{100.00} & 91.98 & \textbf{93.97} & 79.35 & 80.51 \\
 \midrule
\addlinespace[2pt]
ByT5-small & FOD & 79.35 & 82.18 & 74.13 & 71.19 & 76.17 \\
 & Selected TGR & 90.77 & 88.71 & 84.73 & 79.35 & 65.82 \\
 & Oracle TGR & \textbf{100.00} & \textbf{96.59} & 89.94 & 79.14 & 64.78 \\
 \midrule
\addlinespace[2pt]
CodeT5-base & FOD & 76.82 & 82.76 & 78.82 & \textbf{85.81} & 83.98 \\
 & Selected TGR & 87.88 & 92.78 & 83.97 & 82.17 & 81.74 \\
 & Oracle TGR & 98.65 & 93.31 & 91.51 & 85.77 & \textbf{85.03} \\
 \midrule
\addlinespace[2pt]
CodeT5-small & FOD & 79.96 & 78.94 & 79.21 & 76.34 & 78.54 \\
 & Selected TGR & 87.07 & 85.58 & 84.73 & 79.63 & 74.61 \\
 & Oracle TGR & \textbf{100.00} & 90.55 & 90.75 & 83.75 & 75.60 \\
 \midrule
 \midrule
\addlinespace[2pt]
Mean & FOD & 77.63 & 82.77 & 75.39 & 77.53 & \textbf{76.98} \\
 & Selected TGR & 89.25 & 89.44 & 83.74 & 79.50 & 73.43 \\
 & Oracle TGR & \textbf{99.73} & \textbf{92.81} & \textbf{91.24} & \textbf{81.69} & 74.43 \\
\bottomrule
\end{tabular}
\end{table}

\clearpage

\begin{table}[H]
\centering
\normalsize
\setlength{\tabcolsep}{4pt}
\ContinuedFloat
\caption[]{Complete-response fidelity by reference-gap length (continued).}
\begin{tabular}{@{}llrrrrr@{}}
\toprule
& & \multicolumn{5}{c}{Template gap length (characters)} \\
\cmidrule(lr){3-7}
Model & Condition & 0 & 1--16 & 17--64 & 65--256 & 257+ \\
\midrule
\multicolumn{7}{@{}l}{\textit{ROUGE-L F1}} \\
Cowrie & Full response & 53.09 & 65.75 & 42.39 & 19.23 & 18.31 \\
\addlinespace[2pt]
GPT-4o-mini & Full response & 32.67 & 53.50 & 34.69 & 33.12 & 29.33 \\
\midrule
\midrule
\addlinespace[2pt]
Qwen3-0.6B & FOD & 75.97 & 86.49 & 68.24 & 77.14 & 73.31 \\
 & Selected TGR & 92.21 & 92.88 & 85.89 & 85.35 & 71.69 \\
 & Oracle TGR & \textbf{100.00} & 95.03 & 93.82 & 86.93 & 72.77 \\
 \midrule
\addlinespace[2pt]
ByT5-base & FOD & 82.78 & 89.37 & 81.78 & 87.46 & 82.09 \\
 & Selected TGR & 91.61 & 95.03 & 88.87 & 86.05 & 86.36 \\
 & Oracle TGR & \textbf{100.00} & 95.54 & 95.44 & 87.28 & 86.50 \\
 \midrule
\addlinespace[2pt]
ByT5-small & FOD & 82.11 & 85.43 & 77.64 & 78.97 & 80.09 \\
 & Selected TGR & 92.24 & 93.38 & 88.35 & 87.23 & 73.91 \\
 & Oracle TGR & \textbf{100.00} & \textbf{98.42} & 92.67 & 87.04 & 73.99 \\
 \midrule
\addlinespace[2pt]
CodeT5-base & FOD & 78.42 & 86.90 & 81.82 & \textbf{91.59} & 85.61 \\
 & Selected TGR & 90.06 & 95.85 & 87.89 & 88.98 & 86.22 \\
 & Oracle TGR & 99.05 & 95.72 & \textbf{94.40} & 91.44 & \textbf{88.39} \\
 \midrule
\addlinespace[2pt]
CodeT5-small & FOD & 82.48 & 82.63 & 81.86 & 84.96 & 82.36 \\
 & Selected TGR & 89.21 & 90.86 & 88.34 & 87.74 & 80.52 \\
 & Oracle TGR & \textbf{100.00} & 94.64 & 93.49 & 90.21 & 81.55 \\
 \midrule
 \midrule
\addlinespace[2pt]
Mean & FOD & 80.35 & 86.16 & 78.27 & 84.02 & \textbf{80.69} \\
 & Selected TGR & 91.07 & 93.60 & 87.87 & 87.07 & 79.74 \\
 & Oracle TGR & \textbf{99.81} & \textbf{95.87} & \textbf{93.96} & \textbf{88.58} & 80.64 \\
\bottomrule
\end{tabular}
\end{table}

\clearpage
\subsubsection{Complete-Response Length Curves}
\begin{figure}[H]
\centering
\begin{subfigure}[t]{.48\linewidth}
\centering
\includegraphics[width=\linewidth]{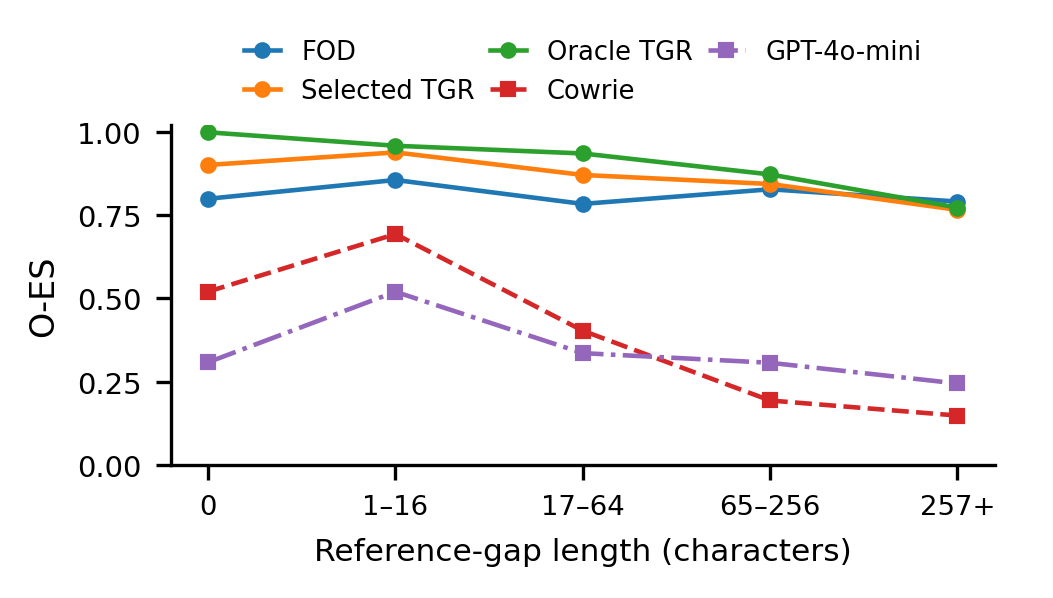}
\caption{Output edit similarity}
\end{subfigure}\hfill
\begin{subfigure}[t]{.48\linewidth}
\centering
\includegraphics[width=\linewidth]{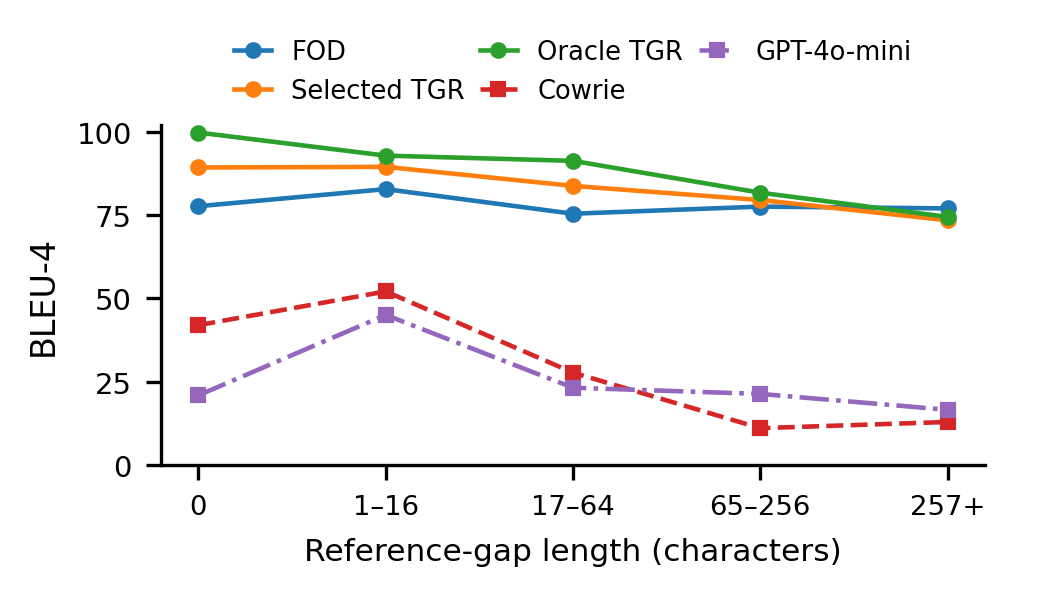}
\caption{Mean BLEU-4}
\end{subfigure}
\par\smallskip
\begin{subfigure}[t]{.48\linewidth}
\centering
\includegraphics[width=\linewidth]{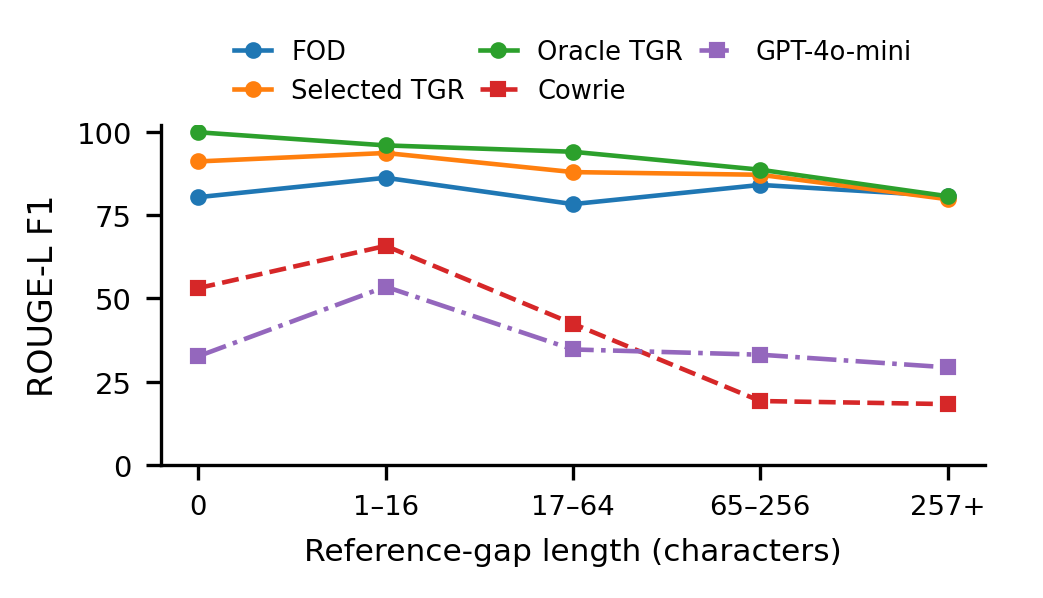}
\caption{ROUGE-L F1}
\end{subfigure}
\caption{Complete-response fidelity for five-models means and both baselines by reference-gap length.}
\label{fig:residual-length-all}
\end{figure}

\clearpage
\section{Gap Prediction Fidelity}
\label{sec:append-gap-results}
\subsection{Gap-Scoring Protocol}
\label{sec:append-gap-protocol}
\label{sec:append-gap-fidelity}
Reference-informed assignment and earliest successive literal matches fix the reference gaps, with predictions being parsed from the saved sentinel sequence.
Template literals and sentinels are not scored, and all expected gaps remain, including missing trailing predictions.
\begin{equation}
\label{eq:append-gap-edit-similarity}
\operatorname{Gap\text{-}ES}(s,\widehat s)
=1-\frac{\sum_i d_{\mathrm{Lev}}(s_i,\widehat s_i)}
{\max(1,\sum_i\max(|s_i|,|\widehat s_i|))}.
\end{equation}
BLEU-4 accumulates counts and lengths within corresponding gap slots of one record before scoring and averaging.
Counts are scored independently within each slot, and neither matches nor candidate n-grams cross a gap boundary.
Empty gap slots receive no separate scores or weights.
Gap ROUGE-L sums the within-gap subsequence lengths and token totals before calculating each record's F1. Those record scores are then averaged.
No n-gram or subsequence match crosses a gap boundary, and no perfect scores are added for individual empty slots.

\subsection{Overall Scores}

\begin{table}[H]
\centering
\normalsize
\caption{Overall oracle-TGR gap fidelity.}
\label{tab:all-gap-results}
\begin{tabular}{@{}lrrr@{}}
\toprule
Model & Gap ES & BLEU-4 & ROUGE-L F1 \\
\midrule
Qwen3-0.6B & 0.597 & 55.91 & 61.67 \\
ByT5-base & 0.614 & 56.11 & 63.61 \\
ByT5-small & 0.601 & 54.84 & 62.57 \\
CodeT5-base & \textbf{0.684} & \textbf{65.76} & \textbf{70.56} \\
CodeT5-small & 0.683 & 63.54 & 69.34 \\
\midrule
Mean & 0.636 & 59.23 & 65.55 \\
\bottomrule
\end{tabular}
\end{table}

\clearpage
\subsection{By Template Coverage}

\begin{table}[H]
\centering
\normalsize
\setlength{\tabcolsep}{5pt}
\caption{Oracle-TGR gap fidelity by reference-informed template coverage. Mean rows average separately calculated model scores.}
\label{tab:gap-coverage-all}
\label{tab:gap-coverage-mean}
\label{tab:gap-coverage-qwen3}
\label{tab:gap-coverage-byt5_base}
\label{tab:gap-coverage-byt5_small}
\label{tab:gap-coverage-codet5_base}
\label{tab:gap-coverage-codet5_small}
\begin{tabular}{@{}lrrrr@{}}
\toprule
& \multicolumn{4}{c}{Template coverage (\%)} \\
\cmidrule(lr){2-5}
Model & $(0,25]$ & $(25,50]$ & $(50,75]$ & $(75,100)$ \\
\midrule
\multicolumn{5}{@{}l}{\textit{Gap ES (0--1)}} \\
Qwen3-0.6B & 0.433 & 0.875 & 0.804 & 0.728 \\
ByT5-base & 0.443 & 0.869 & 0.808 & 0.781 \\
ByT5-small & 0.409 & 0.838 & 0.824 & 0.806 \\
CodeT5-base & \textbf{0.546} & 0.857 & 0.848 & 0.827 \\
CodeT5-small & 0.531 & \textbf{0.885} & \textbf{0.857} & \textbf{0.84}1 \\
\midrule
\addlinespace[2pt]
Mean & 0.472 & 0.865 & 0.828 & 0.797 \\
\midrule
\multicolumn{5}{@{}l}{\textit{BLEU-4 (0--100)}} \\
Qwen3-0.6B & 42.01 & 84.45 & 70.88 & 67.34 \\
ByT5-base & 40.41 & 84.24 & 70.68 & 72.76 \\
ByT5-small & 36.73 & 80.42 & 72.70 & 75.83 \\
CodeT5-base & \textbf{53.97} & 83.53 & \textbf{78.62} & \textbf{77.83} \\
CodeT5-small & 51.79 & \textbf{87.20} & 73.58 & 75.76 \\
\midrule
\addlinespace[2pt]
Mean & 44.98 & 83.97 & 73.29 & 73.90 \\
\midrule
\multicolumn{5}{@{}l}{\textit{ROUGE-L F1 (0--100)}} \\
Qwen3-0.6B & 47.59 & 88.07 & 77.50 & 73.63 \\
ByT5-base & 49.92 & 86.66 & 77.48 & 77.64 \\
ByT5-small & 46.56 & 84.32 & 79.49 & 80.41 \\
CodeT5-base & \textbf{58.88} & 87.03 & \textbf{84.41} & \textbf{82.00} \\
CodeT5-small & 57.16 & \textbf{89.70} & 82.74 & 80.93 \\
\midrule
\addlinespace[2pt]
Mean & 52.02 & 87.16 & 80.32 & 78.92 \\
\bottomrule
\end{tabular}
\end{table}

\subsection{By Reference-Gap Length}
\label{sec:append-gap-length}

\begin{table}[H]
\centering
\normalsize
\setlength{\tabcolsep}{5pt}
\caption{Oracle-TGR gap-only fidelity by reference-gap length. Mean rows average separately calculated model scores.}
\label{tab:gap-length-all}
\begin{tabular}{@{}lrrrr@{}}
\toprule
& \multicolumn{4}{c}{Reference-gap length (characters)} \\
\cmidrule(lr){2-5}
Model & 1--16 & 17--64 & 65--256 & 257--1,024 \\
\midrule
\multicolumn{5}{@{}l}{\textit{Gap ES (0--1)}} \\
Qwen3-0.6B & 0.829 & 0.708 & 0.660 & 0.436 \\
ByT5-base & 0.863 & 0.769 & 0.625 & 0.547 \\
ByT5-small & \textbf{0.878} & 0.747 & 0.687 & 0.472 \\
CodeT5-base & 0.856 & 0.790 & 0.760 & \textbf{0.589} \\
CodeT5-small & 0.876 & \textbf{0.857} & \textbf{0.765} & 0.505 \\
\midrule
Mean & 0.860 & 0.774 & 0.699 & 0.510 \\
\midrule
\multicolumn{5}{@{}l}{\textit{BLEU-4 (0--100)}} \\
Qwen3-0.6B & 80.45 & 68.05 & 56.34 & 43.71 \\
ByT5-base & 84.19 & 75.12 & 51.65 & 53.95 \\
ByT5-small & \textbf{86.07} & 71.59 & 58.29 & 45.47 \\
CodeT5-base & 84.01 & 76.71 & \textbf{69.20} & \textbf{60.00} \\
CodeT5-small & 81.83 & \textbf{83.00} & 65.20 & 49.94 \\
\midrule
Mean & 83.31 & 74.89 & 60.14 & 50.61 \\
\midrule
\multicolumn{5}{@{}l}{\textit{ROUGE-L F1 (0--100)}} \\
Qwen3-0.6B & 82.85 & 71.81 & 66.51 & 47.59 \\
ByT5-base & 86.12 & 77.14 & 62.08 & 58.31 \\
ByT5-small & \textbf{87.76} & 74.24 & 69.32 & 52.20 \\
CodeT5-base & 85.82 & 79.40 & \textbf{77.22} & \textbf{63.12} \\
CodeT5-small & 84.16 & \textbf{85.87} & 76.55 & 54.15 \\
\midrule
Mean & 85.34 & 77.69 & 70.33 & 55.07 \\
\bottomrule
\end{tabular}
\end{table}